\documentclass[sigconf]{acmart}

\usepackage{algorithm}
\usepackage{algpseudocode}
\usepackage{subcaption}
\usepackage{siunitx}

\algdef{SE}[PARFOR]{ParFor}{EndParFor}[1]{\textbf{parfor} \(\mbox{#1}\) \textbf{do}}{\textbf{end parfor}}

\AtBeginDocument{%
  }

\copyrightyear{2024}
\acmYear{2024}
\setcopyright{acmlicensed}\acmConference[GECCO '24]{Genetic and Evolutionary Computation Conference}{July 14--18, 2024}{Melbourne, VIC, Australia}
\acmBooktitle{Genetic and Evolutionary Computation Conference (GECCO '24), July 14--18, 2024, Melbourne, VIC, Australia}
\acmDOI{10.1145/3638529.3654223}
\acmISBN{979-8-4007-0494-9/24/07}

\begin{document}

\title{GPU-accelerated Evolutionary Multiobjective Optimization Using Tensorized RVEA}

\author{Zhenyu Liang}
\orcid{0009-0003-8876-8638}
\email{zhenyuliang97@gmail.com}
\affiliation{%
	\institution{Southern University of Science and Technology}
	\streetaddress{1088 Xueyuan Avenue}
	\city{Shenzhen}
	\state{Guangdong}
	\country{China}
	\postcode{518055}
}

\author{Tao Jiang}
\orcid{0000-0002-0526-4182}
\email{jiangt_97@163.com}
\affiliation{%
	\institution{Southern University of Science and Technology}
    \institution{Peng Cheng Laboratory}
	\streetaddress{1088 Xueyuan Avenue}
	\city{Shenzhen}
	\state{Guangdong}
	\country{China}
	\postcode{518055}
}

\author{Kebin Sun}
\orcid{0009-0008-9213-7835}
\email{sunkebin.cn@gmail.com}
\affiliation{%
	\institution{Southern University of Science and Technology}
	\streetaddress{1088 Xueyuan Avenue}
	\city{Shenzhen}
	\state{Guangdong}
	\country{China}
	\postcode{518055}
}

\author{Ran Cheng}
\orcid{0000-0001-9410-8263}
\authornote{Corresponding Author}
\email{ranchengcn@gmail.com}
\affiliation{%
	\institution{Southern University of Science and Technology}
	\streetaddress{1088 Xueyuan Avenue}
	\city{Shenzhen}
	\state{Guangdong}
	\country{China}
	\postcode{518055}
}

\renewcommand{\shortauthors}{Zhenyu Liang, Tao Jiang, Kebin Sun and Ran Cheng}

\begin{abstract}
	Evolutionary multiobjective optimization has witnessed remarkable progress during the past decades. 
	However, existing algorithms often encounter computational challenges in large-scale scenarios, primarily attributed to the absence of hardware acceleration. 
	In response, we introduce a Tensorized Reference Vector Guided Evolutionary Algorithm (TensorRVEA) for harnessing the advancements of GPU acceleration. 
	In TensorRVEA, the key data structures and operators are fully transformed into tensor forms for leveraging GPU-based parallel computing. 
	In numerical benchmark tests involving large-scale populations and problem dimensions, TensorRVEA consistently demonstrates high computational performance, achieving up to over 1000$\times$ speedups. 
	Then, we applied TensorRVEA to the domain of multiobjective neuroevolution for addressing complex challenges in robotic control tasks. 
    Furthermore, we assessed TensorRVEA's extensibility by altering several tensorized reproduction operators.
	Experimental results demonstrate promising scalability and robustness of TensorRVEA.
	Source codes are available at \url{https://github.com/EMI-Group/tensorrvea}.
\end{abstract}

\ccsdesc[500]{Theory of computation~Evolutionary algorithms}
\ccsdesc[500]{Applied computing~Multi-criterion optimization and decision-making}

\keywords{Evolutionary Multiobjective Optimization, GPU Acceleration, Neuroevolution}

\maketitle

\section{Introduction}
In real-world scenarios, most optimization problems, such as water distribution system management optimization \cite{water_system}, power system planning \cite{power_system}, and DNA sequence design \cite{Sharma2022}, are intrinsically multiobjective optimization problems (MOPs). 
The MOPs present multiple conflicting objectives, thus making it impossible to find a single solution that optimizes all objectives simultaneously. 
Nevertheless, it is feasible to identify a set of trade-off solutions, termed Pareto-optimal solutions.
Formally, a MOP can be formulated as:
\begin{equation}
	\begin{aligned}
		 & \underset{\mathbf{x}}{\text{minimize}} &  & \boldsymbol{f}(\mathbf{x}) = (f_1(\mathbf{x}), f_2(\mathbf{x}), \ldots, f_m(\mathbf{x})), \\
	\end{aligned}
\end{equation}
where  $\mathbf{x} = [x_1, x_2, \ldots, x_d]^{\top} \in \mathbb{R}^d$ is the decision vector with $d$ being the number of decision variables. Meanwhile, $\boldsymbol{f} \in \mathbb{R}^m$ represents the objective vector with \( m \) indicating the number of objectives. It is worth noting that the Pareto-optimal solutions are referred to as the Pareto Set (PS) in the decision space and the Pareto Front (PF) in the objective space. Specifically, problems with $d \geq 100$ are termed Large-Scale Multiobjective Optimization Problems (LSMOPs).

Evolutionary Algorithms (EAs), especially in the domain of Evolutionary Multiobjective Optimization (EMO), have emerged as pivotal in resolving complex MOPs.  
This evolutionary trajectory began with David Goldberg's seminal work in 1989 \cite{goldberg1989}, leading to significant advancements in EMO algorithms since then.
With the ongoing development of contemporary science and engineering, real-world scenarios frequently involve optimization challenges on a large scale, i.e., the LSMOPs \cite{lsmo}.
These problems are distinguished by their complexity, high-dimensional nature, as well as many optimization objectives. 
However, traditional EMO algorithms are designed for CPUs, which leads to substantial processing times, thereby limiting the computational efficiency and scalability of the EMO algorithms. 
Although some algorithms, such as the Reference Vector Guided Evolutionary Algorithm (RVEA) \cite{rvea}, have been conceived with parallelization in mind, the effective processing of LSMOPs still remains challenging due to the absence of hardware acceleration.

GPUs exhibit advanced computational capabilities in many fields \cite{gpu_computing}, primarily due to their outstanding parallel processing abilities. 
In particular, the JAX \cite{jax} framework has been instrumental in advancing GPU-accelerated libraries such as EvoJAX \cite{evojax}, evosax \cite{evosax}, and EvoX \cite{evox}, which enhance evolutionary computation and offer performance improvements over traditional CPU-based methods. 
These developments are crucial for addressing computational challenges in EMO, especially in LSMOPs.
However, the application of GPU acceleration in EMO algorithms has predominantly been concentrated on enhancing the NSGA-II, with other EAs receiving far less focus \cite{nsga2gpu_2,nsga2gpu_3}.

To utilize the advancements of GPU computing, \emph{tensorization} \cite{tensorEC} provides a solution, i.e., transforming the data structures and operators of EMO algorithms into full tensors. 
This is particularly facilitated by GPUs equipped with specialized tensor cores, which enable efficient parallel processing in computation.
The integration of GPU-accelerated computing and tensorization offers a promising avenue for handling the computational demands when solving LSMOPs. 
Nonetheless, research on integrating GPU acceleration and tensorization is limited, existing studies primarily concentrate on low-dimensional numerical optimization \cite{nsga2gpu}, with a significant lack of investigation into high-dimensional, real-world applications.

Consequently, this research aims to combine GPU acceleration techniques and tensorization methods in the realm of EMO. 
To achieve this, we introduce the Tensorized RVEA (TensorRVEA) algorithm.
TensorRVEA is specifically designed to leverage the high computational power of GPUs and the efficient handling of large-scale data structures through tensorization. 
Overall, the main contributions are summarized as follows.
\begin{itemize}
	\item  We introduce TensorRVEA with fully tensorized key data structures and operators optimized for GPU acceleration. In benchmark tests involving large populations and large-scale numerical optimization problems, TensorRVEA attains speedups exceeding 1528$\times$ and 1042$\times$, respectively.

	\item  We apply TensorRVEA to the domain of neuroevolution for addressing complex challenges in multiobjective robotic control tasks~\footnote{We adapt Brax \cite{brax} as the MOPs.}. Experimental results indicate that TensorRVEA still demonstrates high performance in such real-world applications. 

	\item  We demonstrate the compatibility of TensorRVEA by replacing GA \cite{ga} with various tensor-based reproduction operators, including DE \cite{de}, PSO \cite{pso}, and CSO \cite{cso}. 
    With a flexible framework, TensorRVEA highlights the potential in wider applications. 
\end{itemize}

\section{Background}
In this section, we present a brief overview of GPU-accelerated EMO, multiobjective neuroevolution, as well as the original RVEA.
	
\subsection{GPU-accelerated EMO}

Following years of extensive development, EMO algorithms have evolved into three distinct major categories: dominance-based (e.g., NSGA-II \cite{nsga2}), indicator-based (e.g., IBEA \cite{ibea}), and decomposition-based (e.g., MOEA/D \cite{moead}). 
Correspondingly, in the context of many-objective optimization problems, algorithms like NSGA-III \cite{nsga3}, HypE \cite{HypE}, and RVEA \cite{rvea}, of the three categories, have demonstrated their strengths in handling high-dimensional optimization challenges.

Despite the rapid advancements in EMO algorithms, their integration with hardware acceleration remains a significant research gap. 
Existing GPU-accelerated EMO algorithms, such as those presented by Souza \textit{et al.} \cite{moeadaco} and Aguilar-Rivera \textit{et al.} \cite{nsga2gpu}, have demonstrated effectiveness in handling complex MOPs. 
However, these algorithms are primarily implemented using CUDA \cite{cuda}, posing challenges for beginners due to their complexity and the lack of open-source code availability. 

On the other hand, algorithms like TASE \cite{tensorEA1} and TFPSO \cite{tensorEA2}, which incorporate \emph{tensorization} methods with EMO for LSMOPs, are predominantly CPU-based and developed in MATLAB. 
Such methods fail to use the advancement of GPU acceleration fully. 
The potential of tensorization in the EMO field is still considerably untapped, particularly when compared to its widespread impact in machine learning \cite{SurveyTensorTechniques}. 
Therefore, this disparity highlights a promising path for future research, pointing towards the development of more advanced and efficient optimization solutions that leverage the full capabilities of tensorization and GPU acceleration.

Distinguished from existing methods in the literature, the proposed TensorRVEA uniquely integrates tensorized data structures and operators, all finely optimized for GPU acceleration. 
This synthesis of GPU acceleration with full tensorization marks a significant departure from prior algorithms that either partially utilized GPU acceleration or implemented tensorization without fully leveraging GPU capabilities. 
Such a strategic amalgamation in TensorRVEA leads to notable enhancements in computational efficiency and scalability, especially crucial in addressing LSMOPs. 
Moreover, TensorRVEA's design, focused on maximizing GPU computational power and parallel processing capabilities, not only accelerates the processing speed and efficiency but also improves accessibility and usability. 
This advancement is particularly advantageous for newcomers to the field, lowering barriers to entry and simplifying the engagement with advanced EMO techniques.

\subsection{Multiobjective Neuroevolution}

Neuroevolution, a subset of evolutionary computation, is dedicated to evolving neural network architectures and parameters via EAs \cite{ne}. 
This method has proven effective for complex tasks in areas such as robotics, control systems, and game playing. 
Notably, there are emerging attempts to apply EMO algorithms for multiobjective neuroevolution.
For example, Denysiuk \textit{et al.} \cite{smsemoa} utilized the SMS-EMOA algorithm in neuroevolution for multiobjective Knapsack Problems; Stapleton \textit{et al.} \cite{autovehicle} applied NSGA-II to tackle trajectory prediction in autonomous vehicles. 
Using EMO algorithms for multiobjective neuroevolution offers a promising method to balance network performance, computational efficiency, and robustness.

In response, this study extends the use of TensorRVEA to multiobjective neuroevolution, exploiting its efficiency in solving complex tasks. 
Fusing tensorized data structures and operators on GPU-based infrastructures, TensorRVEA demonstrates high computing performance, which is essential for the intensive computational demands of neuroevolution tasks.

\subsection{RVEA}

The reference vector guided evolutionary algorithm (RVEA) \cite{rvea} is tailored for addressing challenges of many-objective optimization.
During the past years, RVEA has attracted significant interest and presents robust performance across a wide range of applications, such as bionic heat sinks \cite{Bionic_heat},  optimization of material designs~\cite{rvea_material_designs}, arrival flight scheduling \cite{flights_scheduling}, hybrid electric vehicle control~\cite{rvea_electric_vehicle_control}, among many others \cite{rvea_applications_6, rvea_applications_7}.

The efficacy of RVEA can be attributed to its simple yet flexible framework. 
As detailed in Appendix \ref{appendix:rvea}, the main iterative loop of RVEA undergoes three phases: offspring generation, fitness evaluation, and reference vector guided selection.
While the first two phases are common across other evolutionary algorithms, the distinctive feature of RVEA lies in its reference vector guided selection.

\begin{figure}[h]
	\centering
	\includegraphics[width=0.3\textwidth]{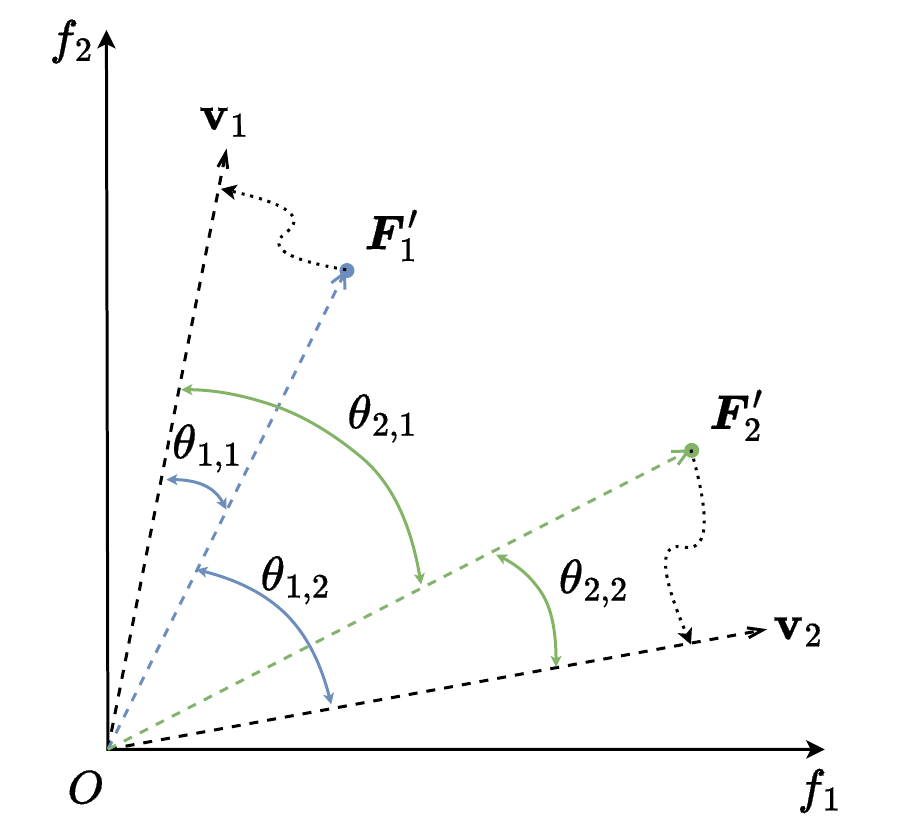}
    \vspace{-5pt}
	\caption{Illustration of reference vector guided selection in RVEA. The population is partitioned into subpopulations by associating each with a reference vector, according to the angle-penalized distance (APD) metric.}
	\label{fig:population_partition}
\end{figure}

As shown in Figure~\ref{fig:population_partition}, as a crucial phase in RVEA, the reference vector guided selection involves partitioning the entire population into subpopulations based on their proximity to the reference vectors in the objective space. 
Subsequently, selection within each subpopulation (or subspace) is independently conducted based on a scalarization measure known as the Angle-Penalized Distance (APD). 
Notably, the tailored design of the selection operation in RVEA is intrinsically ideal for GPU acceleration due to the following reasons.
\begin{itemize}
    \item \textbf{Parallel Processing}: It allows for simultaneous execution of selection operations across different subpopulations, aligning well with the parallel processing capabilities of GPUs.
    \item \textbf{Isolated Computations}: Each subpopulation's selection process, based on the APD metric, involves calculations that are independent of one another, enabling a seamless distribution of tasks across the multiple cores of a GPU without the need for inter-thread communication.
    \item \textbf{Scalability}: The objective space partition mechanism enables the handling of large populations and high-dimensional objective spaces, thus making it well-suited for the scalable computing resources offered by GPUs.
\end{itemize}

\section{Tensorization of RVEA}

Tensorization involves the transformation of traditional data structures and operators into multi-dimensional tensor forms.
In this setting, a tensor is essentially a multi-dimensional array that extends beyond the limitations of matrices, facilitating a more comprehensive and efficient representation of data and computational operators. 
This shift is particularly advantageous for evolutionary algorithms, as it enables the simultaneous processing of multiple solutions, thereby markedly accelerating computational tasks.

In the proposed TensorRVEA, a comprehensive implementation of tensorization is achieved through two principal components: the tensorization of data structures and the tensorization of operators. 
While retaining the overarching framework of the original RVEA, TensorRVEA introduces significant advancements in data handling and operational efficiency. 
Additionally, TensorRVEA is implemented on EvoX \cite{evox}. 
This integration ensures that the enhanced capabilities of tensorization are seamlessly incorporated within the original RVEA structure, thereby optimizing the algorithm's performance in complex optimization scenarios.
The main notations used in this paper are summarized in Table \ref{tab:notations}.

\begin{table}[h]
	\centering
    \captionsetup{skip=5pt}
	\caption{Summary of notations}
	\label{tab:notations}
	\begin{tabular}{ccc}
	\toprule
	Notations & Description &\\
	\midrule
	\( n \) & Population size & \\
	\( d \) & Number of decision variables & \\
	\( m \) & Number of objectives & \\
	\( r \) & Number of reference vectors & \\
	\( \mathbf{x} \) & Decision vector & \\
    \( \boldsymbol{f} \) & Objective vector & \\
	\( \mathbf{v} \) & Reference vector & \\
	\( \boldsymbol{X} \) & Tensorized population  & \\
	\( \boldsymbol{F} \) & Tensorized objective vectors& \\
	\( \boldsymbol{V} \) & Tensorized reference vectors & \\
	\bottomrule
	\end{tabular}
\end{table}

\subsection{Tensorized Data Structures}

In the original RVEA framework, data structures are predominantly set-based. 
By contrast, in TensorRVEA, the crucial data structures including the population, objective vectors, and reference vectors, are transformed into tensor formats. 
Such transformation enables the simultaneous processing of entire populations, which allows for more streamlined and parallelized operations.

\begin{itemize}
    \item \textbf{Tensorized Population}: The population is represented as a tensor $\boldsymbol{X} = \begin{bmatrix}
		\mathbf{x}_1,\mathbf{x}_2,\ldots,\mathbf{x}_n
	\end{bmatrix}^{\top}
	\in \mathbb{R}^{n \times d}$. This tensor format allows for the collective handling of all individuals in the population, thereby streamlining various genetic operators.
    
    \item \textbf{Tensorized Objective Vectors}: Objective vectors are encapsulated in the tensor $\boldsymbol{F} =\begin{bmatrix} \boldsymbol{f}(\mathbf{x}_1), \boldsymbol{f}(\mathbf{x}_2), \ldots,\boldsymbol{f}(\mathbf{x}_n) \end{bmatrix}^{\top} \in \mathbb{R}^{n \times m}$. This representation facilitates efficient evaluation and comparison of individual performances across multiple objectives.
    
    \item \textbf{Tensorized Reference Vectors}: The reference vectors, crucial for guiding the search towards the PF, are structured as $\boldsymbol{V} = \begin{bmatrix} \mathbf{v}_1, \mathbf{v}_2, \ldots,\mathbf{v}_r \end{bmatrix}^{\top} \in \mathbb{R}^{r \times m}$. This tensor format aids in effectively aligning the population with the PF, assisting in the environmental selection process.
\end{itemize}

\subsection{Tensorized Operators}

Tensorization of operators in TensorRVEA involves transforming crucial algorithmic processes into tensor-based formats, including crossover, mutation, and selection. 
Below are the detailed tensor formulations for each operator.

\subsubsection{Tensorized Crossover and Mutation}

In TensorRVEA, the pivotal operations of crossover and mutation are implemented through the Simulated Binary Crossover (SBX) \cite{sbx} and Polynomial Mutation \cite{pm}, respectively. 
These processes are accelerated by employing tensor (matrix) calculations, a method analogous to the techniques used in PlatEMO \cite{platemo1}. 
Specifically, the tensorized SBX and Polynomial Mutation are intricate mathematical operators that benefit from tensor representations for parallel computation. 
This method markedly expedites the evolutionary process.
A comprehensive mathematical exposition of these operations, including detailed formulations and implementation, is available in Appendix \ref{appendix:cross_mut}.
In SBX, TensorRVEA generates offspring through population tensor manipulation for efficient parallel processing. Polynomial Mutation, adapted to tensors, introduces controlled variability into the offspring with specific probabilities and amplitudes.

\begin{algorithm}
    \caption{Selection Operator in TensorRVEA}
    \label{alg:tensorized_rvea_selection}
    \begin{algorithmic}[1]
        \State \textbf{Input:} Population tensor $\boldsymbol{X}$, Objective tensor $\boldsymbol{F}$, Reference vector tensor $\boldsymbol{V}$, Number of generations $t_{\text{max}}$, Current generation $t$ and the rate of change of penalty $\alpha$;
        \State \textbf{Output:} Elite population tensor $\boldsymbol{X}_{\text{elite}}$;

        \State Calculate the minimal objective values $\boldsymbol{z}_t^{*}$;
        \State $\boldsymbol{F}^{\prime} \gets \boldsymbol{F} - \text{repeat}(\boldsymbol{z}_t^{*}, N)$;
		\State $\boldsymbol{\Theta} \gets \arccos\left(\frac{\boldsymbol{F}^{\prime} \cdot \boldsymbol{V}^{\top}}{\lVert\boldsymbol{F}^{\prime} \rVert \cdot \lVert\boldsymbol{V}^{\top}\rVert}\right)$;
		\State $\boldsymbol{A} \gets \text{repeat}(\text{RowMin}{(\boldsymbol{\Theta})}, r)$;

        \State $\boldsymbol{T}_{\text{part}} \gets \text{repeat}(\begin{bmatrix} 0, 1, \ldots, n-1\end{bmatrix}^{\top}, r)$;
        \State $\boldsymbol{I} \gets \text{repeat}(\begin{bmatrix} 0, 1, \ldots, r-1\end{bmatrix}, n)$;
        \State $\boldsymbol{T}_{\text{part}} \gets (1 - |\text{sgn}(\boldsymbol{A} - \boldsymbol{I})|) \odot \boldsymbol{T}_{\text{part}} - |\text{sgn}(\boldsymbol{A} - \boldsymbol{I})|$;
		\State $\boldsymbol{\Gamma} \gets \text{RowMin}{\Bigg(\arccos\left(\frac{\boldsymbol{V} \cdot \boldsymbol{V}^{\top}}{\lVert\boldsymbol{V}\rVert \cdot \lVert\boldsymbol{V}^{\top}\rVert}\right)\Bigg)}$;
        \ParFor{each column $\boldsymbol{t}_{\text{part}}, \boldsymbol{\gamma}, \boldsymbol{\theta}$ in $\boldsymbol{T}_{\text{part}}, \boldsymbol{\Gamma},\boldsymbol{\Theta}$}
            \State $\boldsymbol{T}_{\text{APD}}[:, j] = \Bigg(1 + m \cdot \left(\frac{t}{t_{\text{max}}}\right)^\alpha \cdot \frac{\boldsymbol{\theta}[\boldsymbol{t}_{\text{part}}]}{\boldsymbol{\gamma}}\Bigg) \odot \lVert \boldsymbol{F}^{\prime}[\boldsymbol{t}_{\text{part}}]\rVert$;
        \EndParFor
        \State Replace elements in $\boldsymbol{T}_{\text{APD}}$ with $\texttt{inf}$ where $\boldsymbol{T}_{\text{APD}} = -1$;
        \State $\boldsymbol{I}_{\text{next}} \gets \underset{j=0,1,2,\ldots, r-1}{\text{argmin}}(\boldsymbol{T}_{\text{APD}}[:, j])$;
        \State $\boldsymbol{X}_{\text{elite}} \gets \boldsymbol{X}[\boldsymbol{I}_{\text{next}}]$.

\end{algorithmic}
\end{algorithm}

\subsubsection{Tensorized Selection}
The selection operator plays a significant role in RVEA, balancing convergence and diversity in the high-dimensional objective space during the evolutionary search process. 
Correspondingly, as outlined in Algorithm \ref{alg:tensorized_rvea_selection}, the tensorized selection operator in TensorRVEA comprises three distinct components, which are delineated as follows.

\begin{figure}[h]
	\centering
	\includegraphics[width=0.49\textwidth]{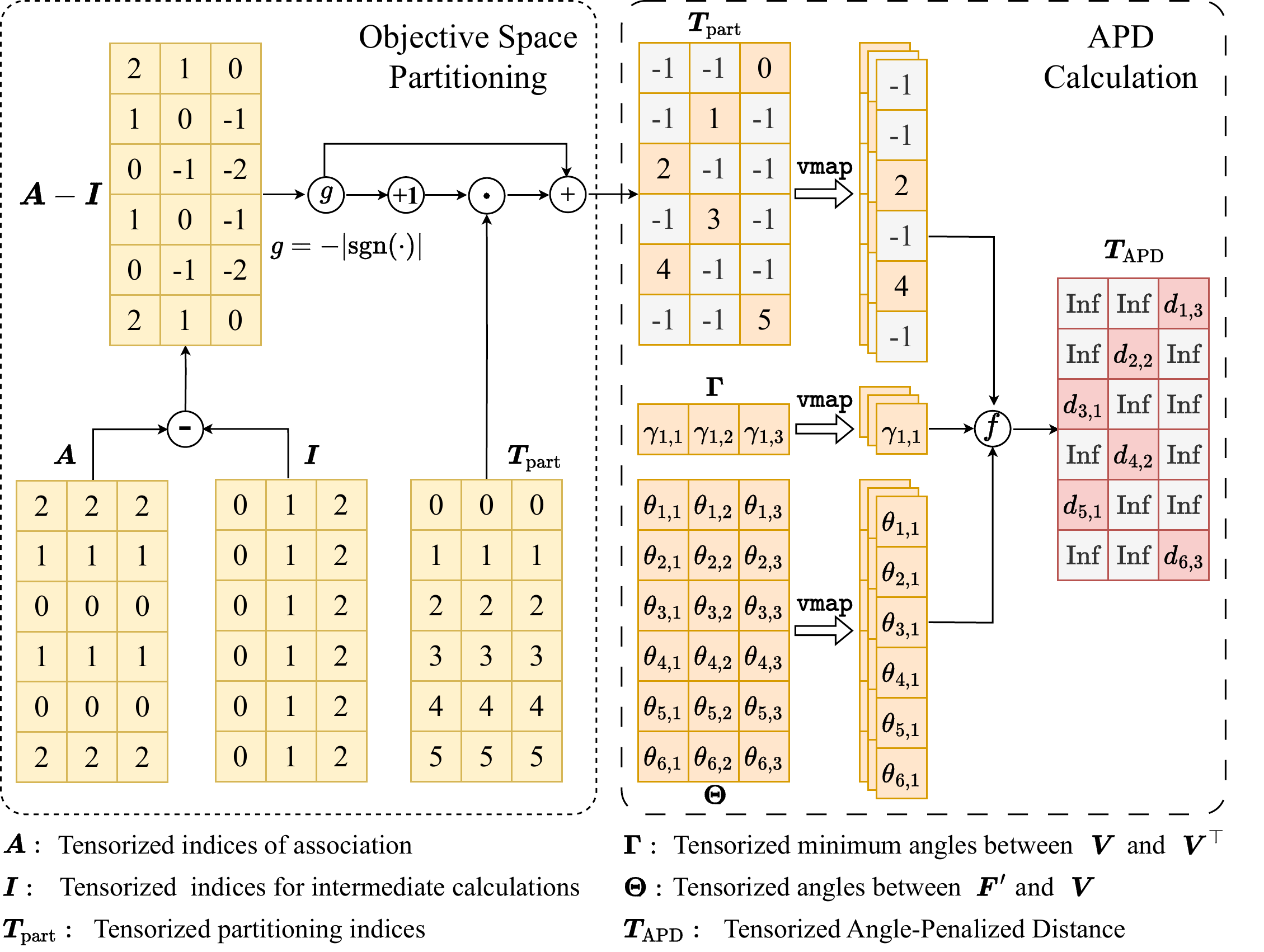}
    \captionsetup{skip=5pt}
	\caption{Example of objective space partitioning and APD calculation for $n=6$, $r=3$. 
    \emph{Left}: objective space partitioning. \emph{Right}: APD calculation.}
	\label{fig:APD}
\end{figure}

\textbf{Translation and Normalization (Lines 3-4)}:
This initial step in the TensorRVEA selection process involves standardizing the objective values across the population to ensure a uniform evaluation basis. 
This standardization is achieved by subtracting the minimal objective values tensor $\mathbf{z}_t^{*}$ from the current objective tensor $\boldsymbol{F}$. 
Executed as a tensor subtraction operation, this process effectively normalizes the objective space, aligning each function's extremum with its respective coordinate axis. 
Such normalization not only facilitates a more equitable comparison of solutions but also sets a crucial foundation for the accurate computation of angles between the objective values and reference vectors in subsequent steps.

\textbf{Objective Space Partitioning (Lines 5-9)}:
In TensorRVEA, the tensorized angle $\boldsymbol{\Theta}$, pivotal in the selection mechanism, quantifies the angles between the translated objective values and the reference vectors, as specified in Line 5 of Algorithm \ref{alg:tensorized_rvea_selection}.
In RVEA, operations are primarily based on vectors. 
By contrast, TensorRVEA enhances this method by transitioning to tensor operations, thereby streamlining the calculation of all angle values in a single and unified process. 
This transformation not only streamlines the computational workflow but also significantly enhances the algorithm's efficiency by processing multiple data points concurrently.

The selection operator then proceeds to calculate the tensor of association  $\boldsymbol{A}$, which plays a crucial role in the population partitioning. 
$\boldsymbol{A}$ is derived by identifying the indices of the smallest angles for each individual to the reference vectors, as obtained from $\boldsymbol{\Theta}$. 
The formulation for $\boldsymbol{A}$ is given by:
\begin{equation}
	\boldsymbol{A} = \text{repeat}(\text{RowMin}(\boldsymbol{\Theta}), r),
\end{equation}
where the $\text{repeat}(\cdot)$ function replicates the column indices of minimal angles, ensuring that each individual is associated with the reference vector most aligned with it. 
$\text{RowMin}(\cdot)$ is a function that identifies the index of the minimum value in each row of the tensorized angle.

Subsequent to the initialization phase, the population in TensorRVEA is partitioned into distinct subpopulations. 
This partitioning is based on proximity metrics derived from the angle calculations. 
While the original RVEA algorithm employs set operations for partitioning the population, TensorRVEA streamlines this process using efficient tensor operations, thus simplifying the procedure. 
As depicted in Figure \ref{fig:APD}, this phase involves constructing a tensor for partitioning indices, denoted as $\boldsymbol{T}_{\text{part}}$, along with a tensor of indices. 
Both tensors have dimensions of $n \times r$. 
The partitioning process is substantially optimized through the following tensor operation:
\begin{equation}
    \boldsymbol{T}_{\text{part}} = (1 - |\text{sgn}(\boldsymbol{A} - \boldsymbol{I})|) \odot \boldsymbol{T}_{\text{part}} - |\text{sgn}(\boldsymbol{A} - \boldsymbol{I})|,
\end{equation}
where $\boldsymbol{I}$ is a tensor of indices with each row ranging from 0 to $r-1$, and $\boldsymbol{T}_{\text{part}}$ is initially populated with column indices ranging from 0 to $n-1$. 
This operation systematically updates the partitioning tensor, assigning indices to individuals closely aligned with the reference vectors, and marking those without close alignment with a value of $-1$.
In Figure \ref{fig:APD}, the value $-1$ is the light gray background, indicating that no individual is associated with this position.

\textbf{APD Calculation and Selection (Lines 10-16)}:
Following RVEA, the selection operator is also based on calculating the Angle-Penalized Distance (APD) for each individual while leveraging vectorizing map (\texttt{vmap}) functions in JAX for efficient parallel computing. 
Specifically, APD is a scalar metric calculated as:
\begin{equation}
	\label{eq:4}
	d_{i, j} = \Bigg(1+m \cdot \left(\frac{t}{t_{max}}\right)^\alpha \cdot \frac{\theta_{i, j}}{\gamma_{\mathbf{v}_{j}}}\Bigg) \cdot \lVert \boldsymbol{f}^{\prime}_{i}\rVert,
\end{equation}
\begin{equation}
	\label{eq:5}
	\gamma_{\mathbf{v}_{j}} = \underset{i \in \{1,2,\ldots, r\}, i \neq j} \min \langle\mathbf{v_{i}}, \mathbf{v_{j}}\rangle,
\end{equation}
where $d_{i, j}$ is the APD of $i$-th individual in the subpopulation corresponding to the $j$-th reference vector, $m$ is the number of objectives, $r$ is the number of reference vectors, $t$ is current generation index,  $t_{max}$ is the maximal number of generations, $\alpha$ is the parameter controlling the rate of change of penalty, $\gamma_{\mathbf{v}_{j}}$ is the smallest angle between reference vector $\mathbf{v}_{j}$ and other reference vectors.

In TensorRVEA, analogous to the computation of $\Theta$, the minimum angles between tensorized reference vectors are determined using tensor operations, replacing the traditional vector-based methods. 
The APD computation, detailed in lines 11-13 of Algorithm \ref{alg:tensorized_rvea_selection}, is executed in parallel on the GPU using the \texttt{parfor} operation.
This parallelism is realized through the \texttt{vmap} function in JAX, facilitating simultaneous processing of loop contents on the GPU and thereby significantly accelerating the operation. 
Illustrated on the right side of Figure \ref{fig:APD}, this procedure involves parallel computation of the APD value for each column of the tensors $\boldsymbol{T}_{\text{part}}, \boldsymbol{\Gamma}, \text{ and } \boldsymbol{\Theta}$. 
These columns are then aggregated, with occurrences of -1 being replaced by \texttt{inf}, leading to the final APD values. 
For each subpopulation, the individual with the smallest APD value is selected as the elite solution.  
This criterion strategically balances convergence towards the PF and diversity within the solution set, thus fostering a comprehensive exploration of the objective space. 
In the case that all APD values in a subpopulation are \texttt{inf}, the individuals are considered invalid.

\section{Experiments}

This study encompasses four comprehensive experiments to rigorously evaluate TensorRVEA: acceleration performance, numerical benchmarks, multiobjective neuroevolution, and experiments demonstrating its extensibility by altering several reproduction operators. 
For fair comparisons, all experiments were consistently conducted on the EvoX platform \cite{evox}, utilizing a system equipped with an AMD EPYC 7543 8-Core Processor server and an NVIDIA® GeForce RTX 4090 GPU. Detailed experimental setups are provided in Appendix \ref{appendix:exp_all}. 

\subsection{Acceleration Performance}

In this experiment, the acceleration performance of TensorRVEA is rigorously compared against the standard RVEA on the DTLZ1 problem \cite{DTLZ}. 
We conducted a comprehensive evaluation involving algorithms: RVEA (CPU), TensorRVEA (CPU), RVEA (GPU), and TensorRVEA (GPU), with the parenthetical notations indicating the computational devices used for each algorithm's execution. 
Two sets of experiments were conducted for each algorithm, each iteration independently repeated 10 times to ensure robustness. 
In the first series, we standardized the decision dimension and the number of objectives at \( d = 100 \) and \( m = 3 \), respectively, while varying the population size from 32 ($2^5$) to \num{16384} ($2^{14}$). 
The algorithms were run over 100 generations, during which we calculated the average computation time per generation. 
The second series maintained a constant population number (\( n = 100 \)) and the number of objectives (\( m = 3 \)), but varied the decision dimension, doubling it from 512 ($2^9$) to \num{262144} ($2^{18}$). 
Again, these algorithms were run for 100 generations to ascertain the average time per generation. 
The primary focus of these experiments was to analyze the speedup achieved by TensorRVEA (GPU) compared to RVEA (CPU), particularly under scenarios of large populations and high-dimensional decision contexts, thus understanding acceleration performance in handling extensive computational loads and large-scale problems.

\begin{figure}[h]
	\centering
	\includegraphics[width=0.38\textwidth]{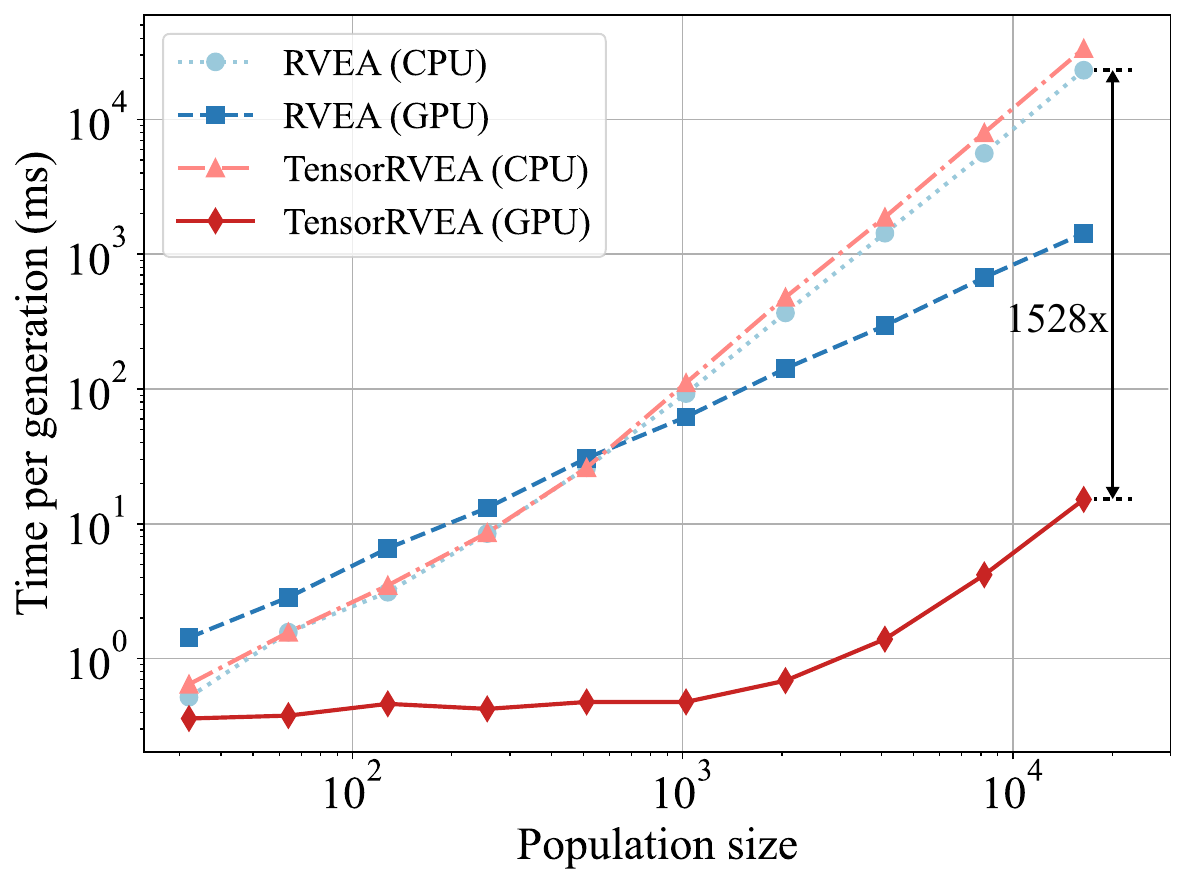}
    \captionsetup{skip=-1pt}
	\caption{Performance comparison on CPU and GPU platforms when scaling the population size on DTLZ1 problem. 
 Results highlight the significant speedup achieved by using GPU over CPU, with TensorRVEA on GPU showing a remarkable \num{1528}$\times$ speedup at the largest population size examined.
 }
	\label{fig:pop_time}
\end{figure}

\begin{figure}[h]
	\centering
	\includegraphics[width=0.38\textwidth]{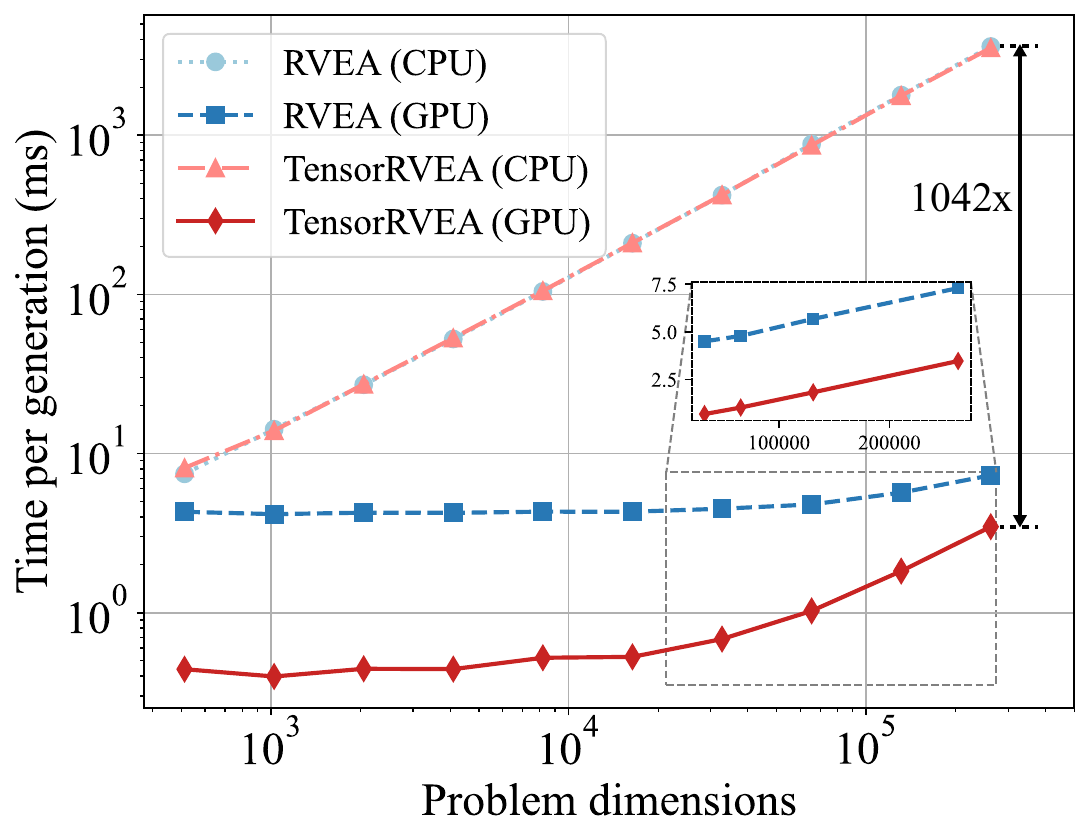}
    \captionsetup{skip=-1pt}
	\caption{
 Performance comparison on CPU and GPU platforms when scaling the problem dimension on DTLZ1 problem. 
 Results highlight the significant speedup achieved by using GPU over CPU, with TensorRVEA on GPU showing a remarkable \num{1042}$\times$ speedup at the largest problem dimension examined.
 }
	\label{fig:dim_time}
\end{figure}

As evidenced in Figure \ref{fig:pop_time}, a discernible trend is apparent in terms of runtime and population for both RVEA (CPU) and TensorRVEA (CPU). 
By contrast, a striking acceleration ratio is evident when comparing TensorRVEA (GPU) to RVEA (CPU), with the former achieving up to 1528$\times$ speedups. 
Moreover, the runtime for TensorRVEA (GPU) demonstrates relative stability across a wide range of population sizes, from 32 to \num{1024}. 

It is posited that beyond a population size of \num{1024}, the hardware may encounter overhead limitations. 
Notably, the performance curve of RVEA (GPU) is observed to intersect with that of the CPU variant. 
This can be attributed to RVEA's implementation using JAX's ``\texttt{fori\_loop}`` function, which may incur consistent overheads.

As evidenced in Figure \ref{fig:dim_time}, with the increasing problem dimensions, there is a corresponding rise in the time per generation for algorithms executed on the CPU.
By contrast, the runtime of TensorRVEA (GPU) consistently remains lower than that of RVEA (GPU), achieving speedups of up to 1042$\times$ compared to RVEA (CPU). 
However, a noticeable increase in runtime is observed as the problem dimension escalates to \num{16384}. 
This surge can likely be attributed to the upper limitation of the GPU's computing capacity reached. 
Upon examining the last four data points on a normal axis, it becomes evident that the slopes are very similar, indicating the consistent scalability of TensorRVEA. 

\subsection{Numerical Benchmarks}

In this experiment, we assess the model performance of TensorRVEA and RVEA in terms of Inverted Generational Distance (IGD) \cite{igd} values on DTLZ1-DTLZ4 numerical optimization problems.
We conducted 31 independent runs with a population size of 105 and the algorithm stop time being \SI{90}{\milli\second}.

\begin{figure}[h]
    \centering

    \begin{subfigure}[b]{0.33\linewidth}
        \includegraphics[width=\linewidth]{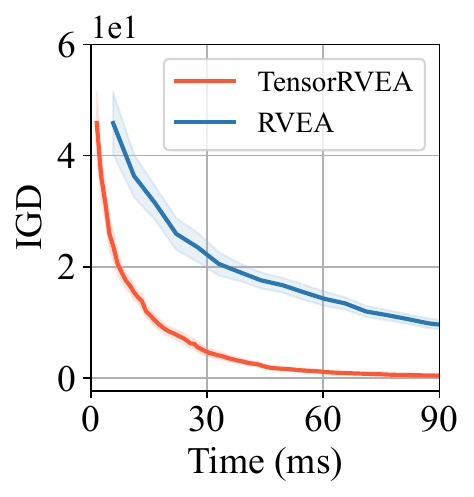}
		\captionsetup{skip=-2pt}
        \caption{DTLZ1}
        \label{fig:DTLZ1}
    \end{subfigure}
    \hspace{5mm}
	\begin{subfigure}[b]{0.33\linewidth}
        \includegraphics[width=\linewidth]{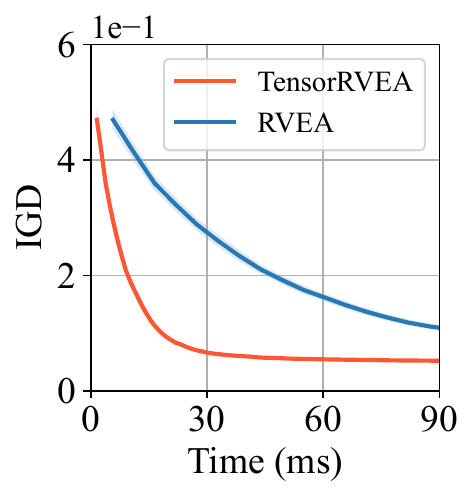}
		\captionsetup{skip=-2pt}
        \caption{DTLZ2}
        \label{fig:DTLZ2}
    \end{subfigure}

    \vspace{1mm}
    
    \begin{subfigure}[b]{0.33\linewidth}
        \includegraphics[width=\linewidth]{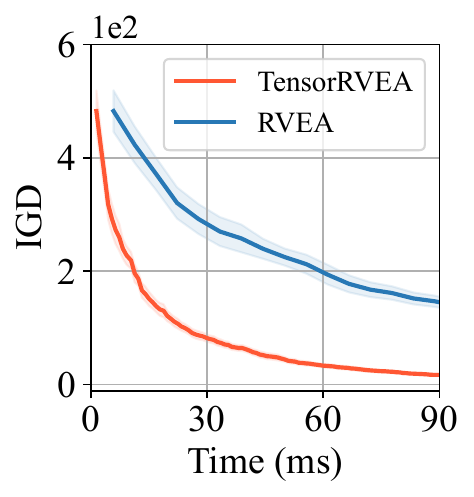}
		\captionsetup{skip=-2pt}
        \caption{DTLZ3}
        \label{fig:DTLZ3}
    \end{subfigure}
    \hspace{5mm}
	\begin{subfigure}[b]{0.33\linewidth}
        \includegraphics[width=\linewidth]{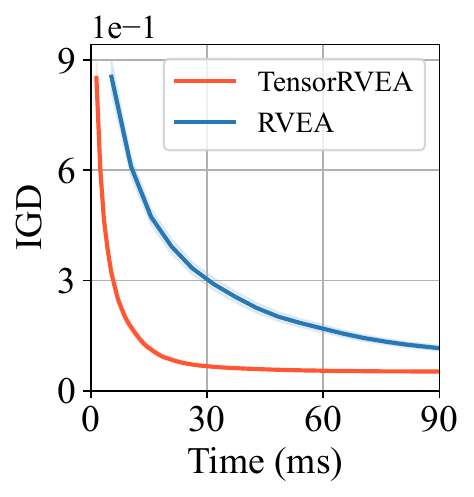}
		\captionsetup{skip=-2pt}
        \caption{DTLZ4}
        \label{fig:DTLZ4}
    \end{subfigure}

    \captionsetup{skip=5pt}
    \setlength{\belowcaptionskip}{-5pt}
    \vspace{-5pt}
    \caption{Comparison of mean IGD values and 95\% confidence intervals for TensorRVEA and RVEA on DTLZ1-DTLZ4 problems.}
    \label{fig:all_DTLZ}
\end{figure}

Figure \ref{fig:all_DTLZ} presents a compelling comparison between TensorRVEA and RVEA. 
The IGD curves for TensorRVEA indicate a rapid convergence within just \SI{90}{\milli\second}, a timeframe in which RVEA has not converged. Remarkably, TensorRVEA nearly reaches 100 generations in this short duration, underlining its exceptional efficiency. 
This performance reveals the benefits of full tensorization.

In all four cases examined, TensorRVEA not only achieves lower IGD values significantly faster than RVEA but also consistently finds solutions closer to the true PF within the same timeframe. 
Most progress towards the PF occurs early in the generation process, particularly in the initial \SI{30}{\milli\second}. 
This is indicated by the rapid improvement in IGD values for both algorithms during these early stages. 
Furthermore, the shaded areas representing the 95\% confidence intervals of the IGD values are noticeably tighter for TensorRVEA than for RVEA, especially on DTLZ1 and DTLZ3. 
It suggests that TensorRVEA outperforms RVEA on average and demonstrates more consistent performance across different scenarios.

\subsection{Multiobjective Neuroevolution}

Brax \cite{brax} is a recently developed physics engine, primarily known for its applications in robotic control. 
However, the original Brax merely provides single-objective problems. 
To conduct the experiment for multiobjective neuroevolution, we drew inspiration from MO-Gymnasium \cite{mogym} and Prediction-Guided MORL \cite{PGMORL}, thereby adapting Brax to facilitate multiobjective optimization problems. 
Our design includes a suite of problems that encompass both two-objective and three-objective settings of several robotic control environments.
For a summary and detailed descriptions of these multiobjective robotic control problems, refer to Table \ref{tab:environment-objectives} and Appendix \ref{appendix:mone}, respectively.
Notably, considering the number of decision variables involved, all problems fall into the scope of LSMOPs.

\begin{table}[htbp]
	\centering
    \captionsetup{skip=5pt}
	\caption{Robotic control tasks for Multiobjective Neuroevolution}
	\label{tab:environment-objectives}
	\resizebox{\columnwidth}{!}{%
	\begin{tabular}{lccc}
		\toprule
		\textbf{Problems} & \textbf{$m$} & \textbf{$d$} & \textbf{Optimization Objectives} \\
		\midrule
		MoHalfCheetah & 2 & 390 & Forward reward, Control cost \\
		MoHopper-m2   & 2 & 243 & Forward reward, Height \\
		MoHopper-m3   & 3 & 243 & Forward reward, Height, Control cost \\
		MoSwimmer     & 2 & 178 & Forward reward, Control cost \\
		\bottomrule
	\end{tabular}}
\end{table}

\begin{figure*}[ht]
    \centering
    \newlength{\figheight}
    \setlength{\figheight}{4.3cm} 

    \newlength{\figheightLastRow}
    \setlength{\figheightLastRow}{4.3cm}

    \begin{subfigure}{.24\linewidth}
        \centering
        \includegraphics[height=\figheight]{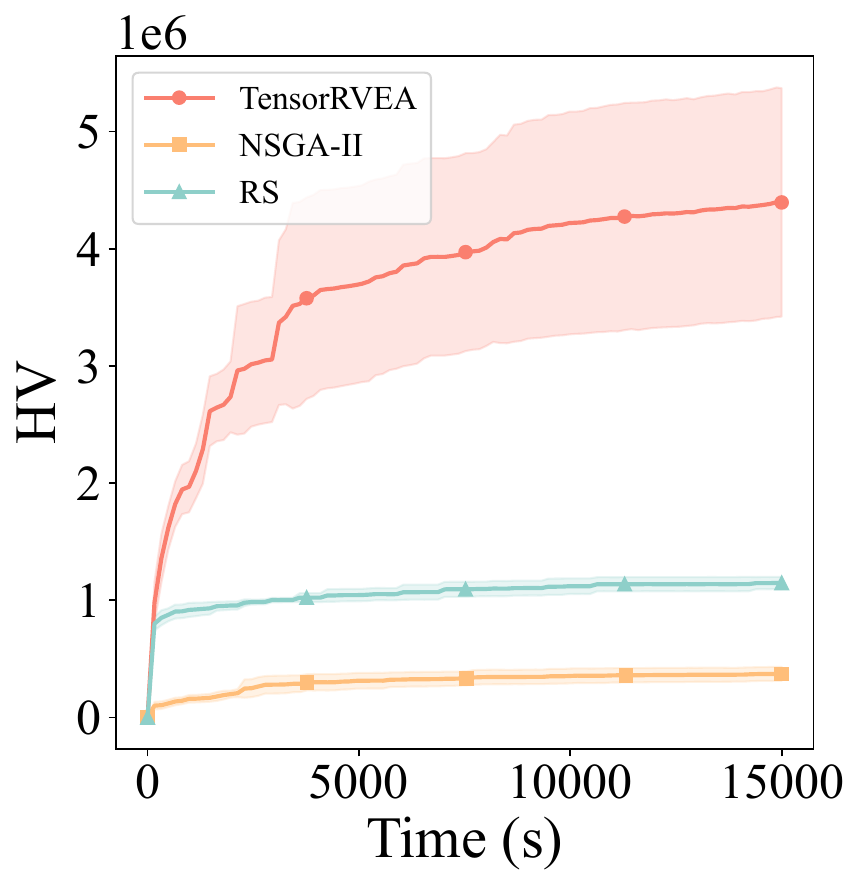}
        \captionsetup{skip=-1pt}
        \caption{}
        \label{fig:hv_halfcheetah}
    \end{subfigure}
    \begin{subfigure}{.24\linewidth}
        \centering
        \includegraphics[height=\figheight]{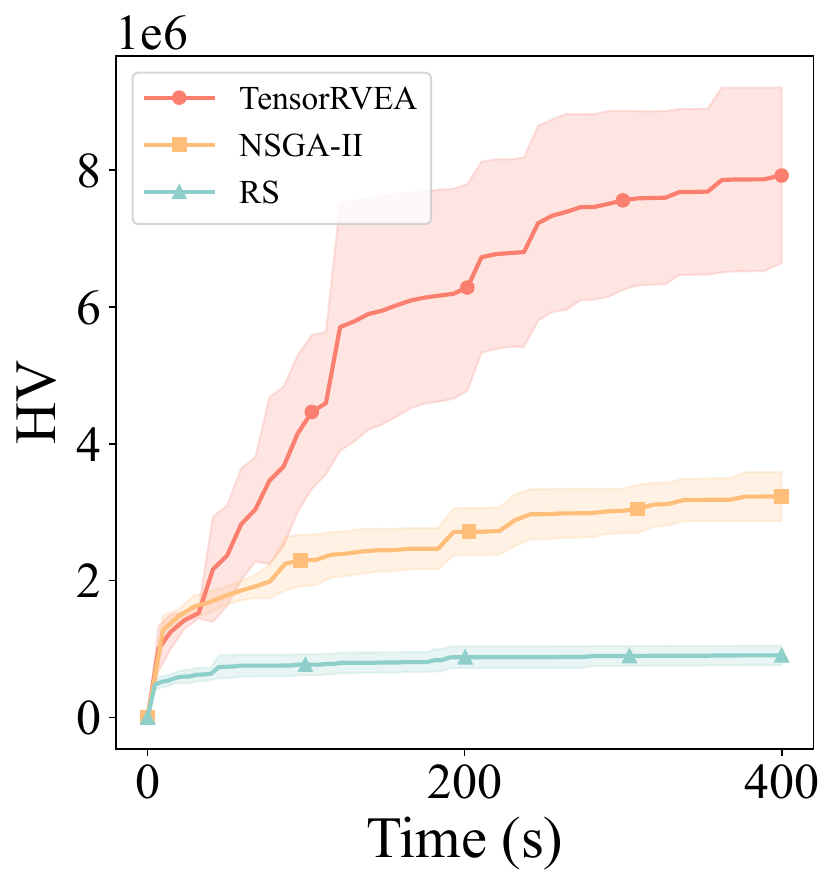}
        \captionsetup{skip=-1pt}
        \caption{}
        \label{fig:hv_hopper}
    \end{subfigure}
    \begin{subfigure}{.24\linewidth}
        \centering
        \includegraphics[height=\figheight]{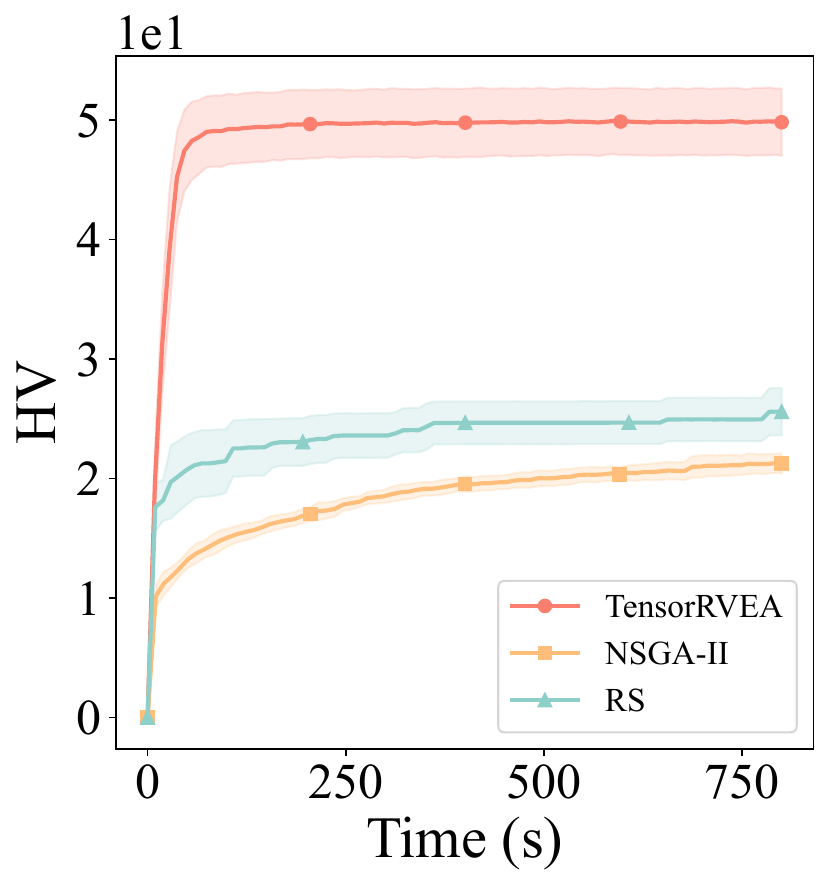}
        \captionsetup{skip=-1pt}
        \caption{}
        \label{fig:hv_swimmer}
    \end{subfigure}
    \begin{subfigure}{.24\linewidth}
        \centering
        \includegraphics[height=\figheight]{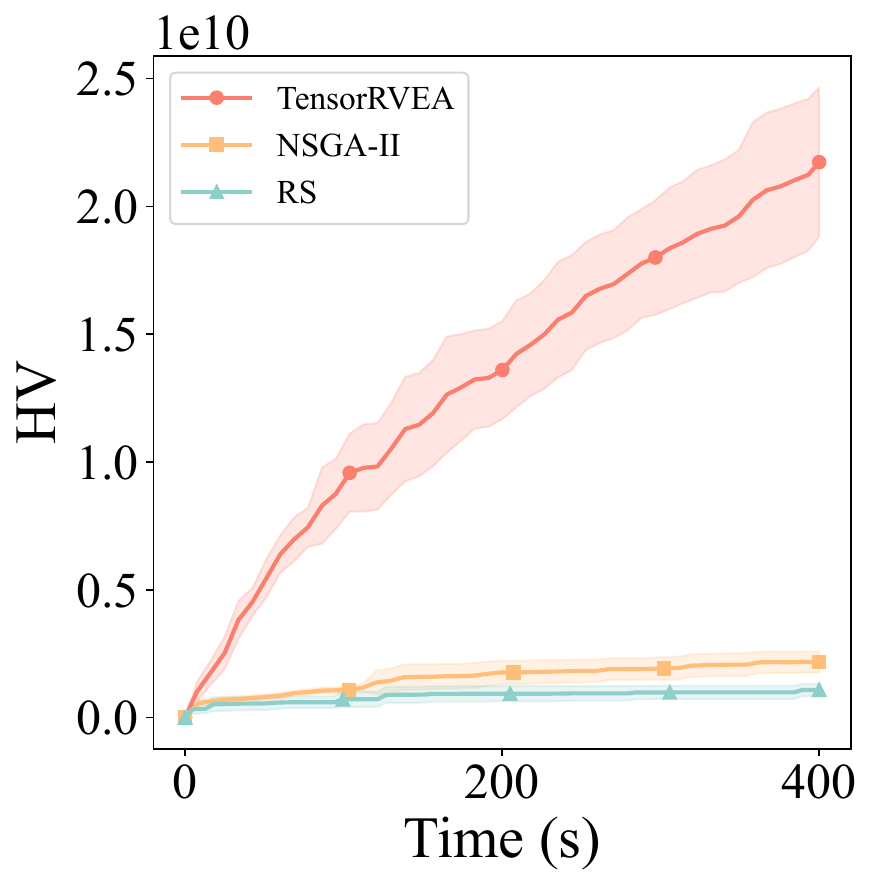}
        \captionsetup{skip=-1pt}
        \caption{}
        \label{fig:hv_hopper_m3}
    \end{subfigure}

    \begin{subfigure}{.24\linewidth}
        \centering
        \includegraphics[height=\figheight]{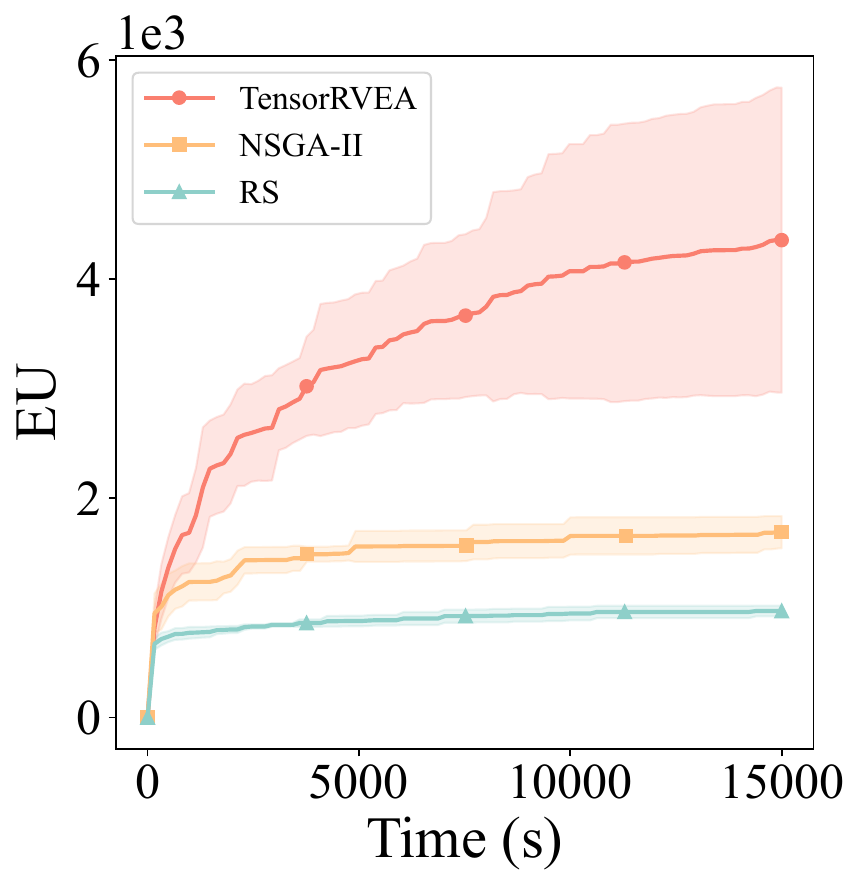}
        \captionsetup{skip=-1pt}
        \caption{}
        \label{fig:eu_halfcheetah}
    \end{subfigure}
    \begin{subfigure}{.24\linewidth}
        \centering
        \includegraphics[height=\figheight]{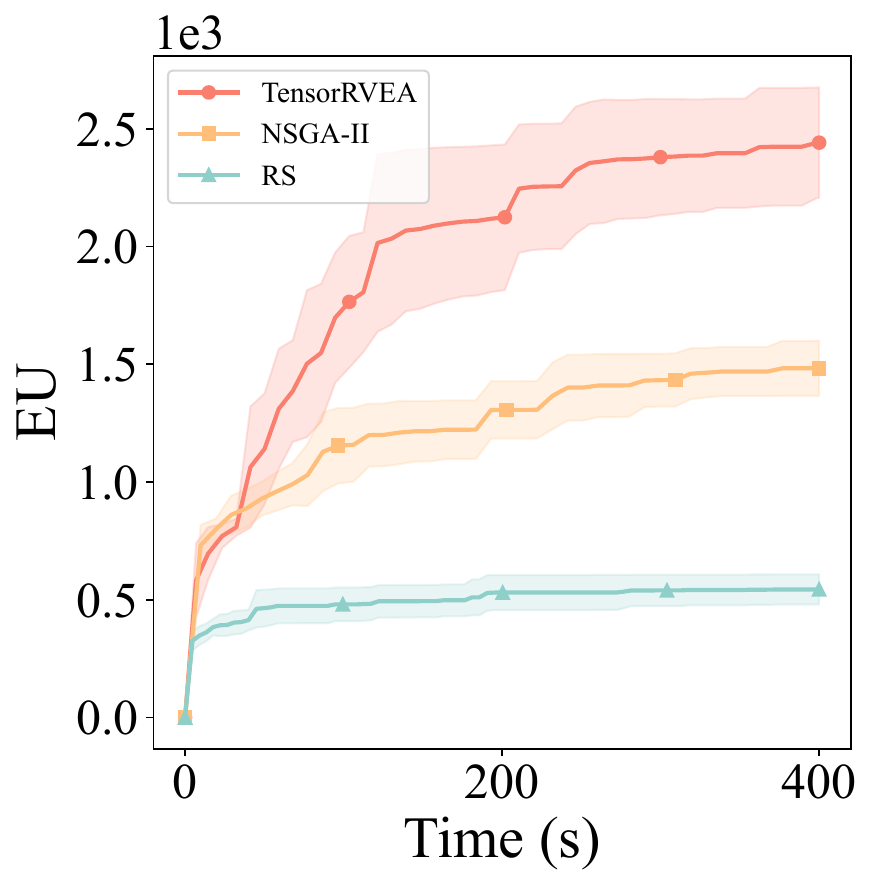}
        \captionsetup{skip=-1pt}
        \caption{}
        \label{fig:eu_hopper}
    \end{subfigure}
    \begin{subfigure}{.24\linewidth}
        \centering
        \includegraphics[height=\figheight]{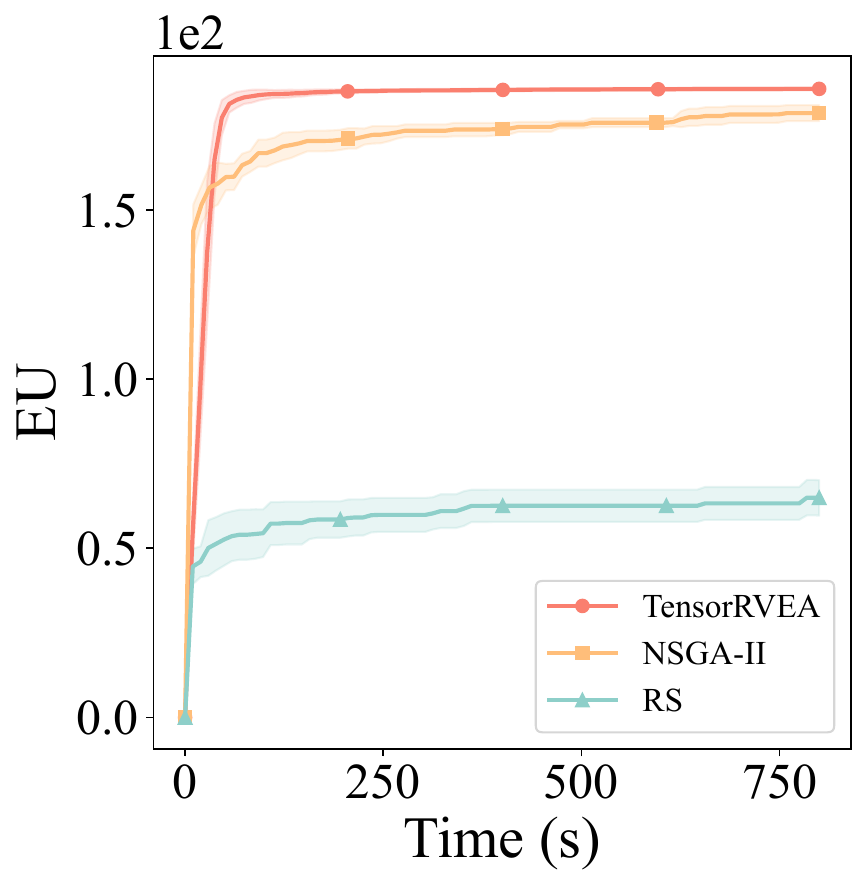}
        \captionsetup{skip=-1pt}
        \caption{}
        \label{fig:eu_swimmer}
    \end{subfigure}
    \begin{subfigure}{.24\linewidth}
        \centering
        \includegraphics[height=\figheight]{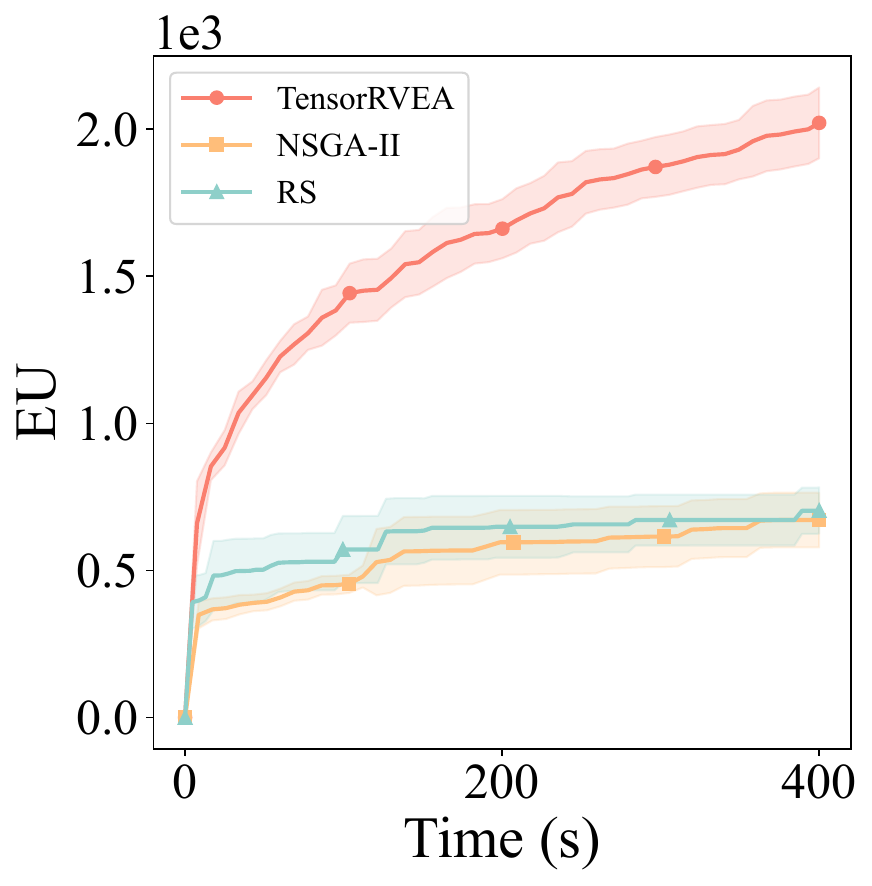}
        \captionsetup{skip=-1pt}
        \caption{}
        \label{fig:eu_hopper_m3}
    \end{subfigure}

    \begin{subfigure}{.24\linewidth}
        \centering
        \includegraphics[height=\figheightLastRow]{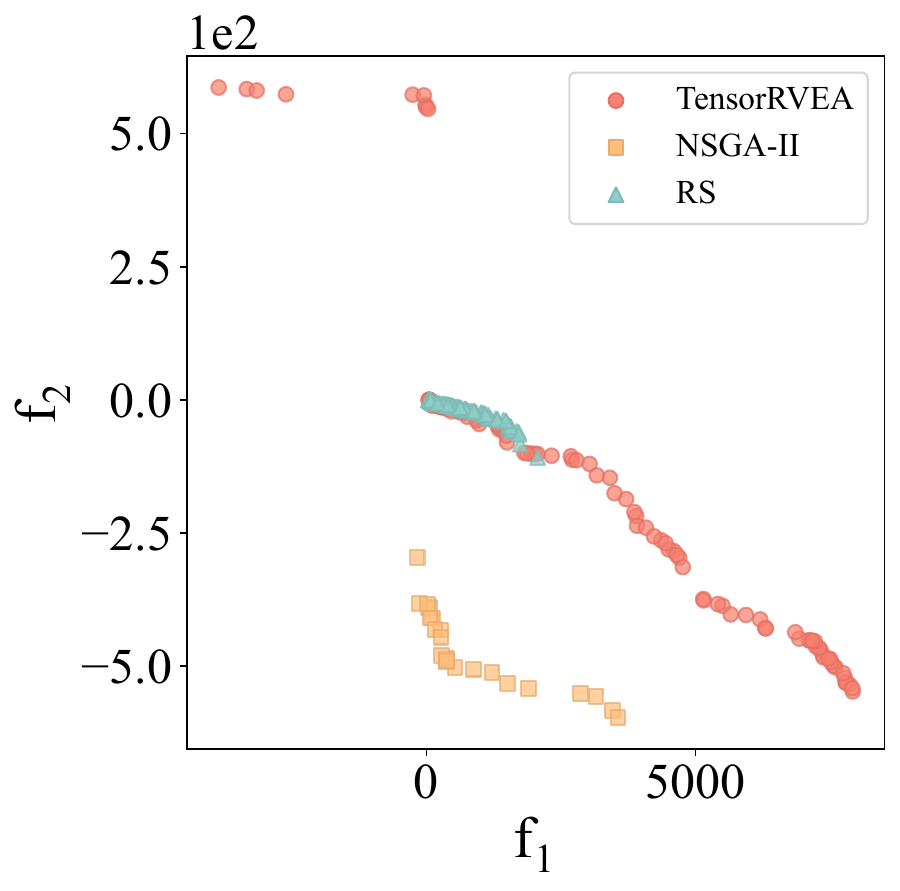}
        \captionsetup{skip=-1pt}
        \caption{}
        \label{fig:pf_halfcheetah}
    \end{subfigure}
    \begin{subfigure}{.24\linewidth}
        \centering
        \includegraphics[height=\figheightLastRow]{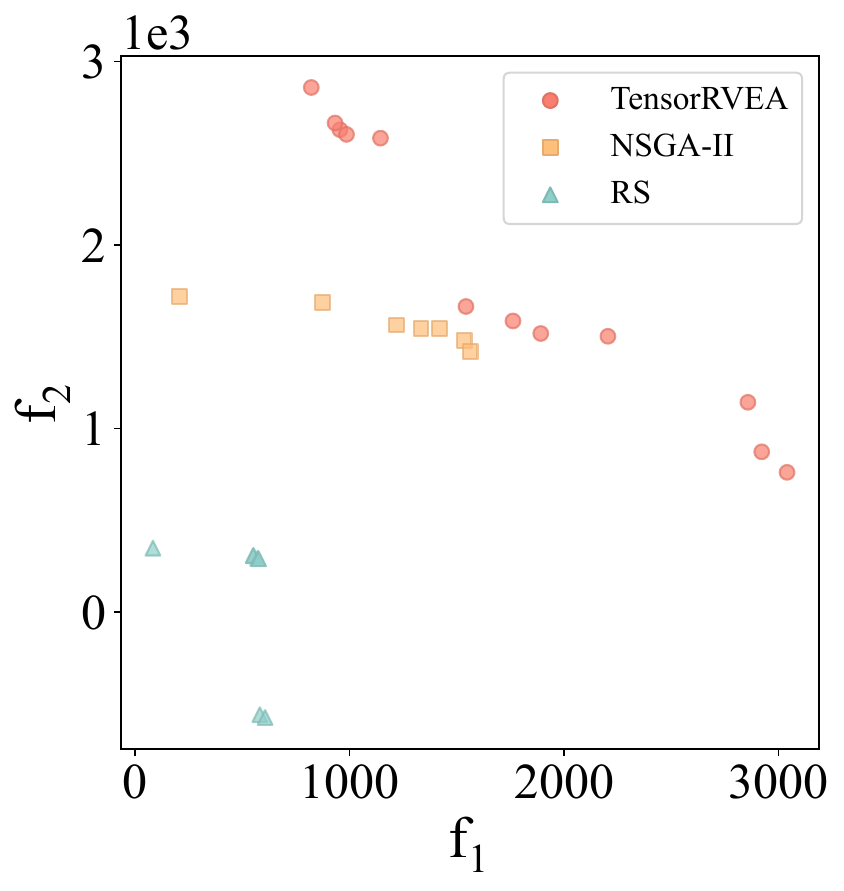}
        \captionsetup{skip=-1pt}
        \caption{}
        \label{fig:pf_hopper}
    \end{subfigure}
    \begin{subfigure}{.24\linewidth}
        \centering
        \includegraphics[height=\figheightLastRow]{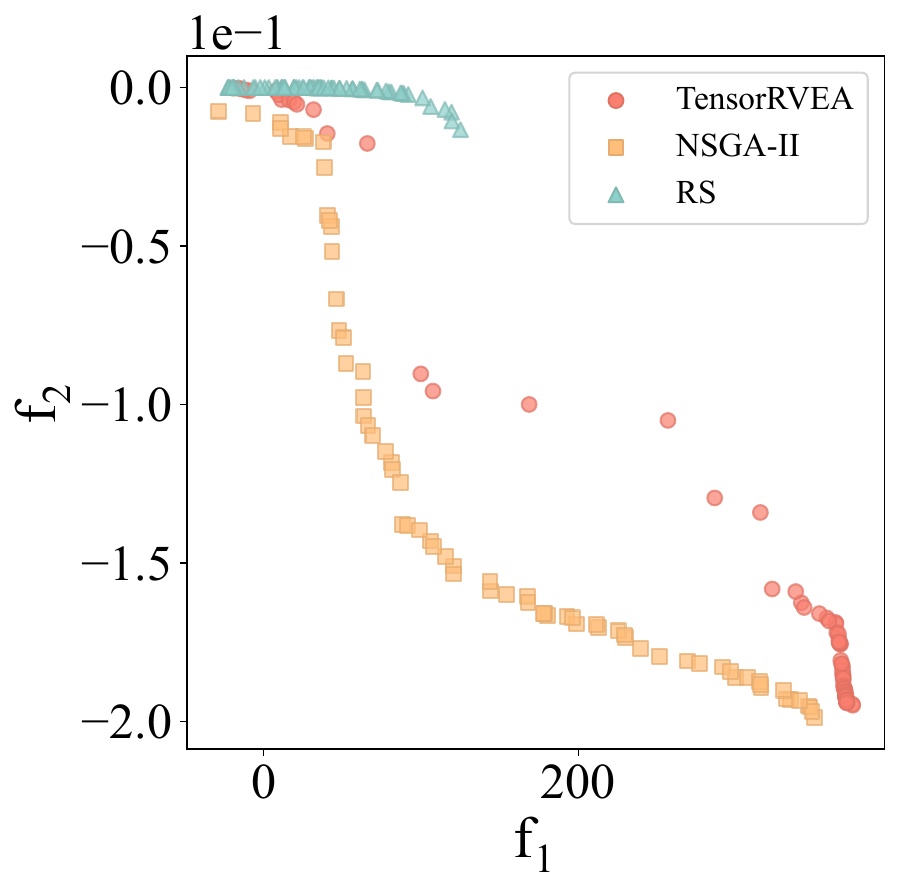}
        \captionsetup{skip=-1pt}
        \caption{}
        \label{fig:pf_swimmer}
    \end{subfigure}
    \begin{subfigure}{.24\linewidth}
        \centering
        \includegraphics[width=4.1cm, height=4.1cm]{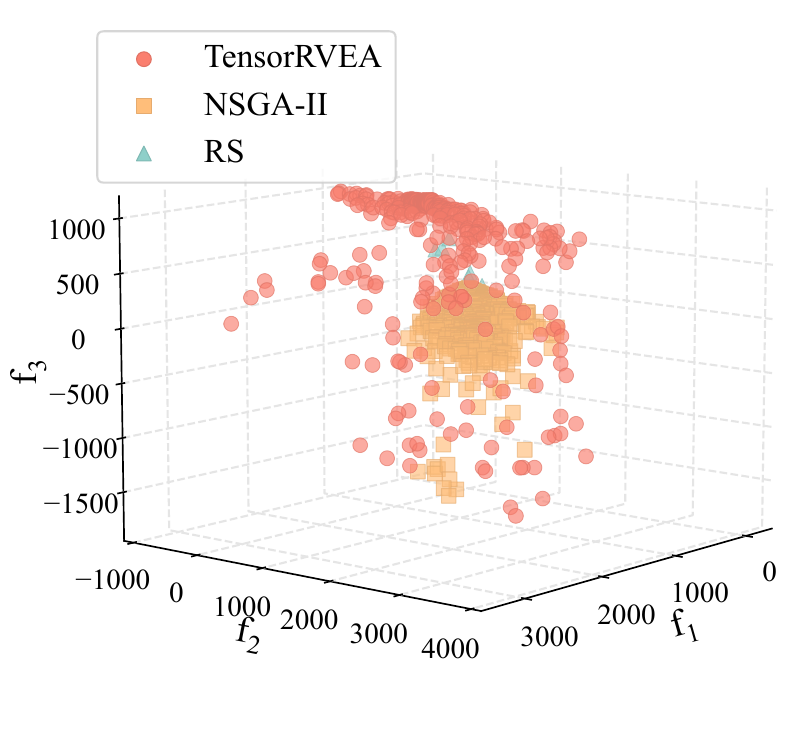}
        \captionsetup{skip=-1pt}
        \caption{}
        \label{fig:pf_hopper_m3}
    \end{subfigure}

    \captionsetup{skip=5pt}
    \caption{Performance comparison of TensorRVEA, NSGA-II, and Random Search (denoted as RS) across MoHalfCheetah, MoHopper-m2, MoSwimmer and MoHopper-m3. Subfigures (a), (b), (c) and (d) depict HV on MoHalfCheetah, MoHopper-m2, MoSwimmer and MoHopper-m3, respectively. Subfigures (e), (f), (g) and (h) depict EU on MoHalfCheetah, MoHopper-m2, MoSwimmer and MoHopper-m3, respectively. Subfigures (i), (j), (k) and (l) show final solution sets on MoHalfCheetah, MoHopper-m2, MoSwimmer and MoHopper-m3, respectively. \emph{Note}: larger values indicate better performances.}
    \label{fig:RL1}
\end{figure*}

\begin{figure*}[t!]
    \centering

    \begin{subfigure}{0.31\linewidth}
        \includegraphics[width=\linewidth]{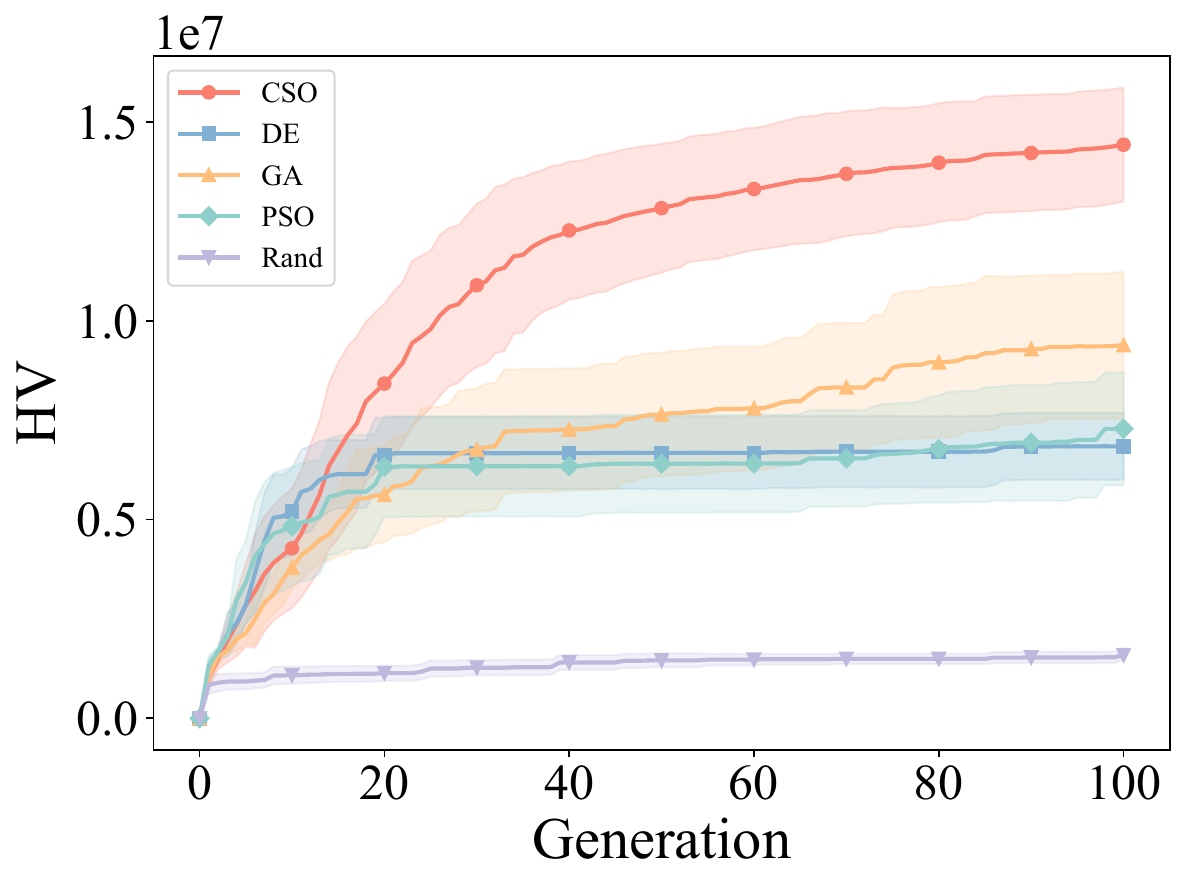}
		\captionsetup{skip=-1pt}
        \caption{HV for MoHopper-m2}
        \label{fig:hv_hopper_m2}
    \end{subfigure}
    \hfill
    \begin{subfigure}{0.31\linewidth}
        \includegraphics[width=\linewidth]{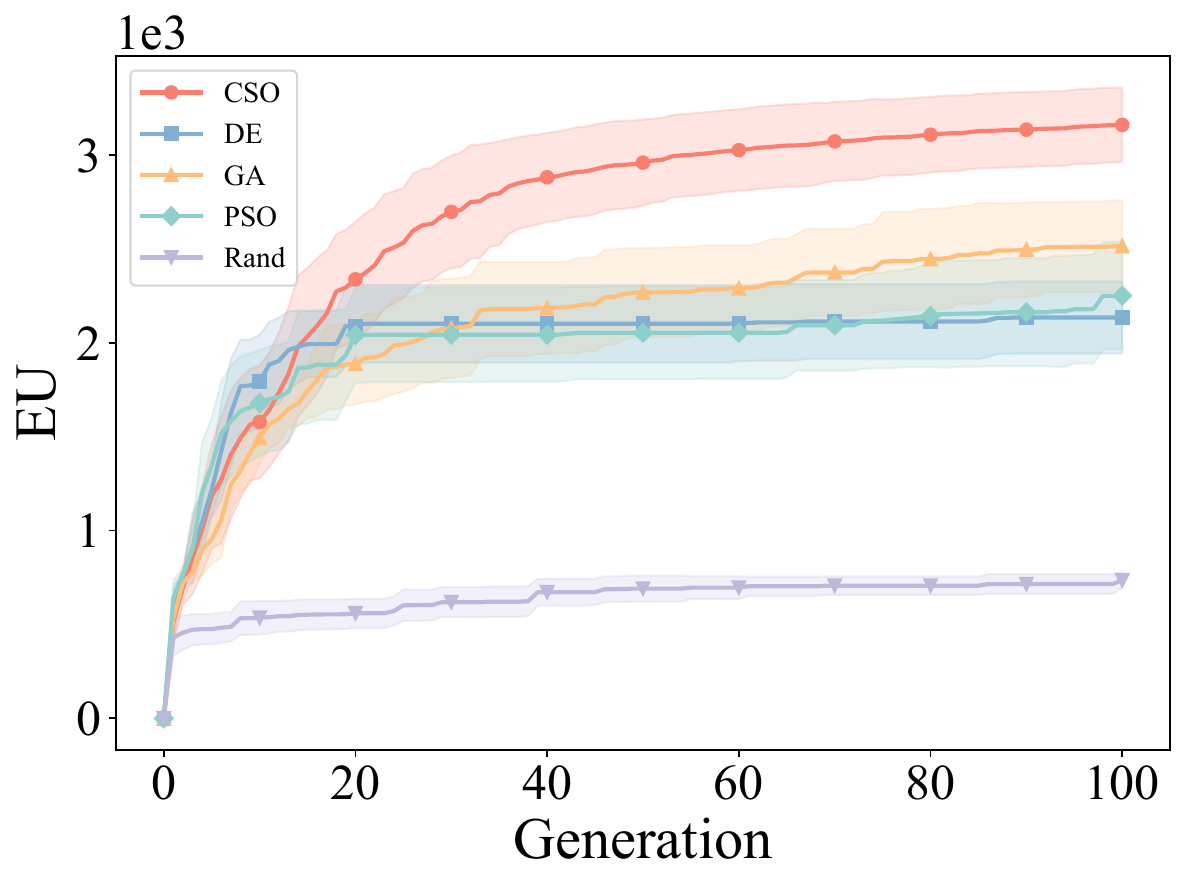}
		\captionsetup{skip=-1pt}
        \caption{EU for MoHopper-m2}
        \label{fig:eu_hopper_m2}
    \end{subfigure}
    \hfill
    \begin{subfigure}{0.31\linewidth}
        \includegraphics[width=\linewidth, height=1.554in]{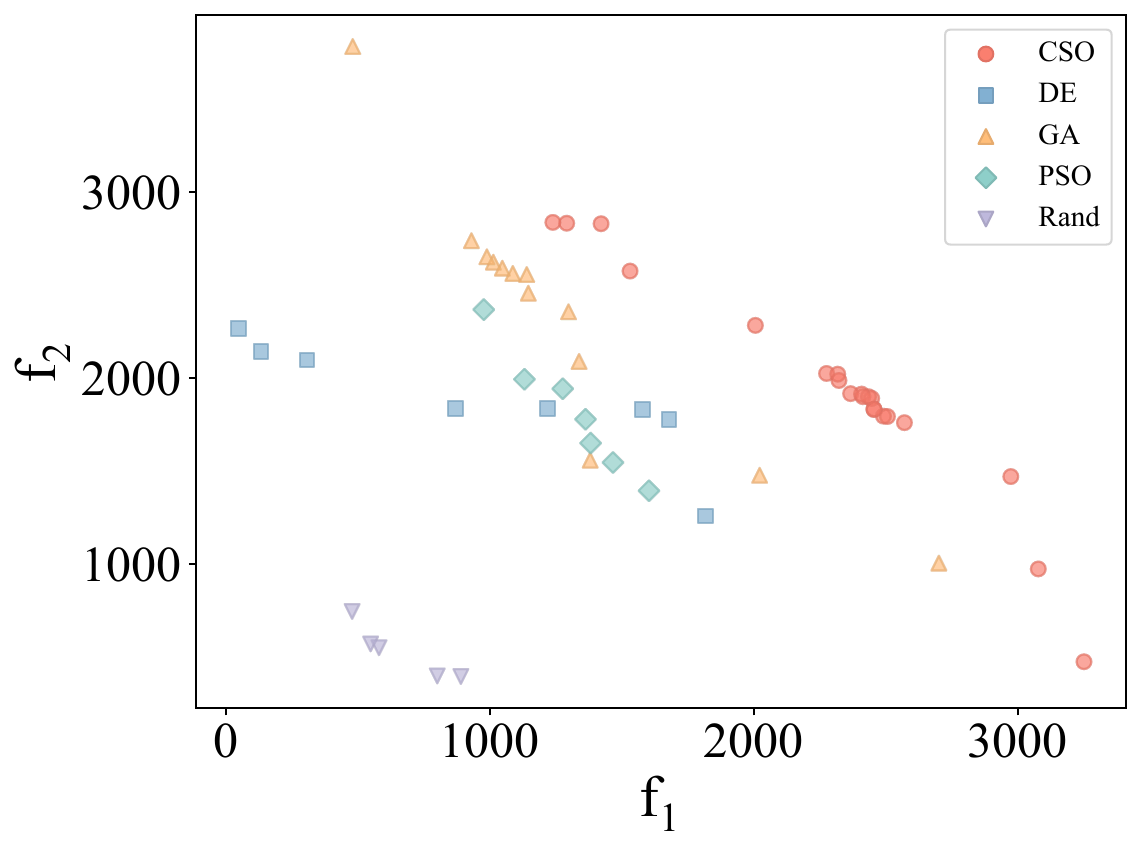}
		\captionsetup{skip=-1pt}
        \caption{Final solution set for MoHopper-m2}
        \label{fig:pf_hopper_m2}
    \end{subfigure}

    \begin{subfigure}{0.31\linewidth}
        \includegraphics[width=\linewidth]{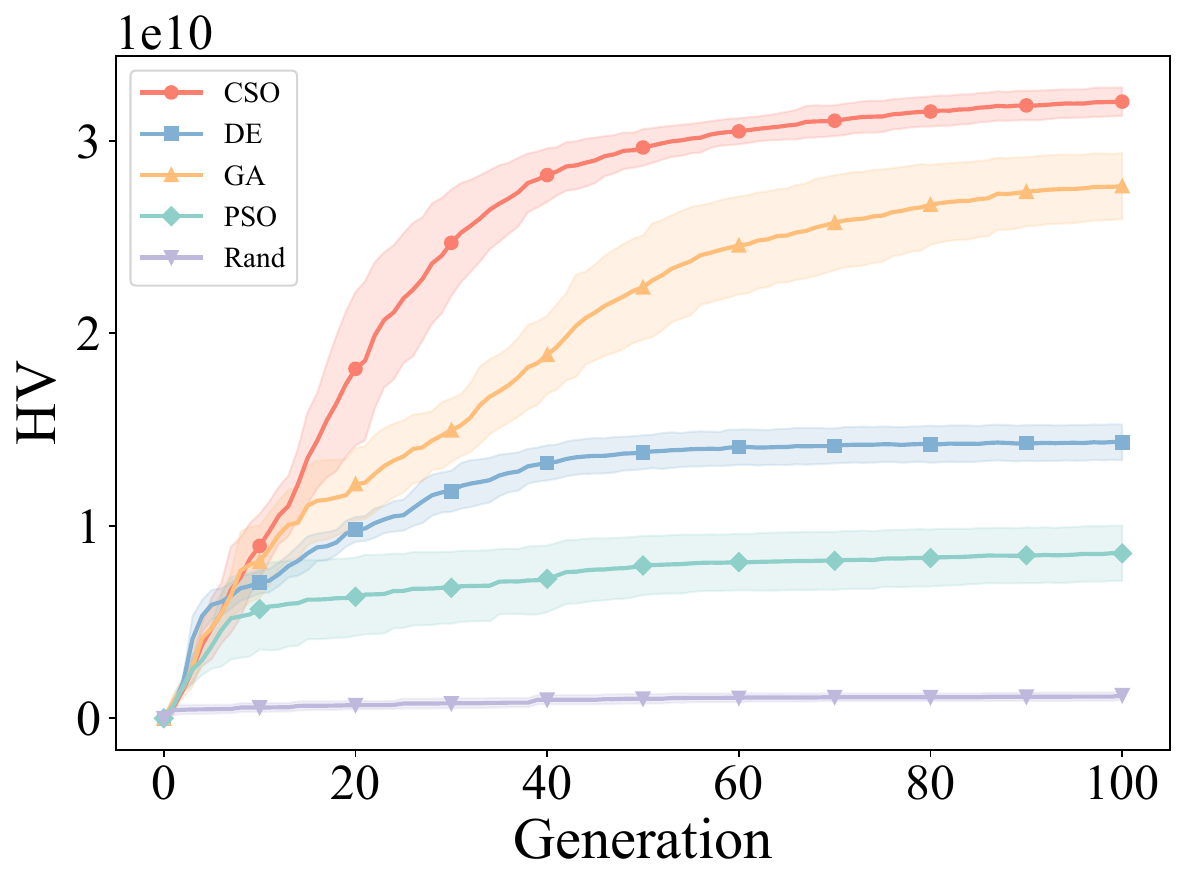}
		\captionsetup{skip=-1pt}
        \caption{HV for MoHopper-m3}
        \label{fig:hv_hopper_m3}
    \end{subfigure}
    \hfill
    \begin{subfigure}{0.31\linewidth}
        \includegraphics[width=\linewidth]{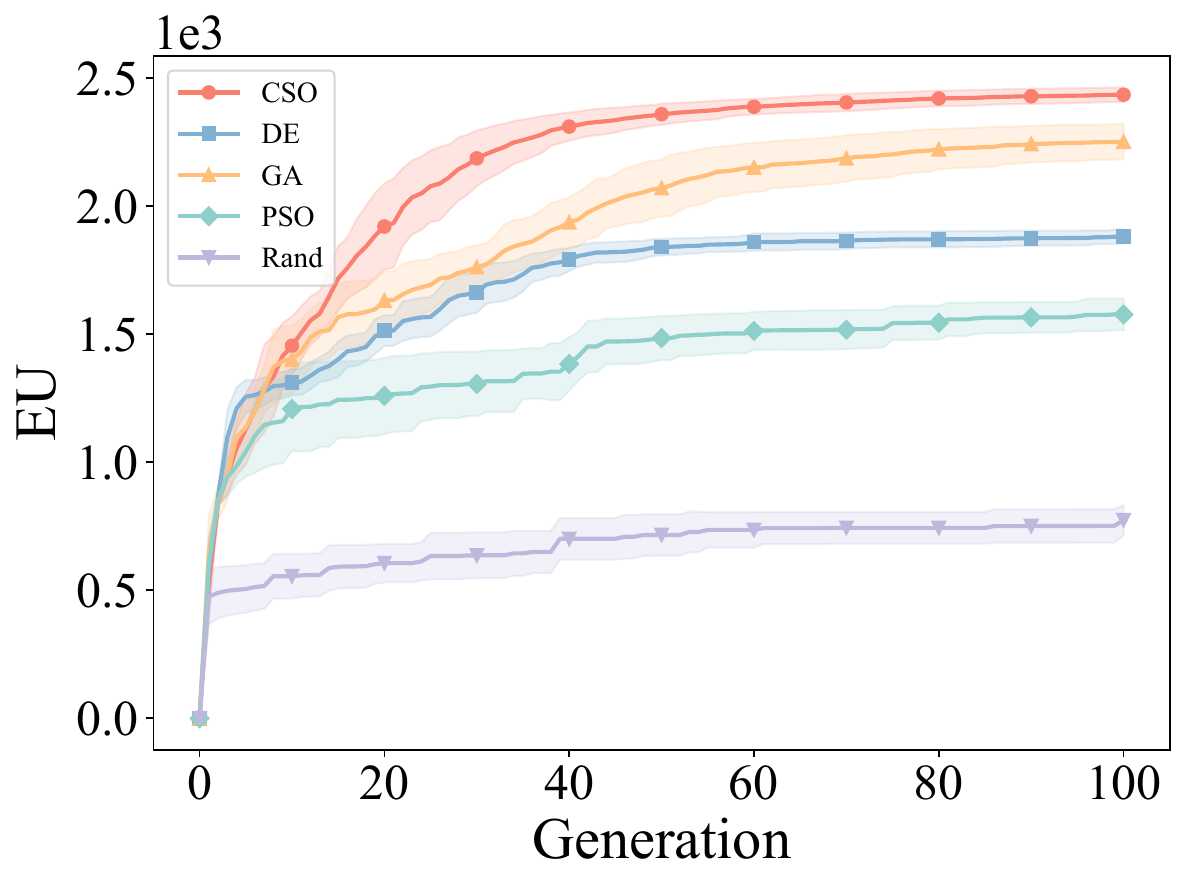}
		\captionsetup{skip=-1pt}
        \caption{EU for MoHopper-m3}
        \label{fig:eu_hopper_m3}
    \end{subfigure}
    \hfill
    \begin{subfigure}{0.31\linewidth}
        \includegraphics[width=\linewidth, height=1.65in]{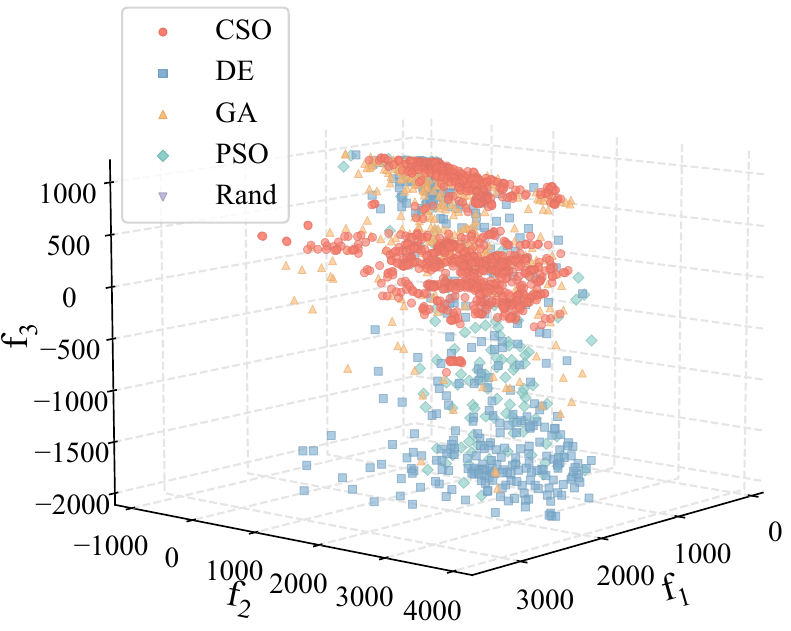} 
		\captionsetup{skip=-1pt}
        \caption{Final solution set for MoHopper-m3}
        \label{fig:pf_hopper_m3}
    \end{subfigure}

    \captionsetup{skip=5pt}
    \caption{Performance comparison of TensorRVEA integrated with CSO, DE, GA, PSO and Random reproduction (denoted as Rand) operators on MoHopper-m2 and MoHopper-m3 problems. \emph{Note}: larger values indicate better performances.}
    \label{fig:RL2}
\end{figure*}

We conducted a comparative analysis of TensorRVEA, tensor-based NSGA-II, and Random Search (RS) across various environments, running each algorithm independently 10 times for statistical robustness. 
Each algorithm was applied to optimize the weights of a Multi-Layer Perceptron (MLP) network, which acts as the policy model, for multiobjective robotic control tasks.
The population of each algorithm was set at \num{10000}. 
Recognizing the complexity of multiobjective robotic control tasks as expensive black-box optimization challenges, we imposed specific time constraints for each test environment: \SI{15000}{\second} for MoHalfCheetah, \SI{400}{\second} for MoHopper-m2, \SI{800}{\second} for MoSwimmer, and \SI{400}{\second} for MoHopper-m3.
During the execution of each algorithm, the non-dominated solutions were recorded at the end of each generation. 
This method was instrumental in capturing the evolutionary progress and identifying the most effective strategies developed over the course of the search. 
The algorithms' performances were meticulously evaluated using the Hypervolume (HV) \cite{hv}, along with the Expected Utility (EU) \cite{eu}, a measure of preference metrics, as well as final solutions visualizations.
The HV metric evaluates the PF's coverage in objective space, reflecting solution diversity and convergence. Conversely, the EU metric gauges solution utility for decision-making, indicating effectiveness in satisfying various criteria.
For detailed definitions, refer to Appendix \ref{appendix:mone}.

As shown in Figure \ref{fig:RL1}, TensorRVEA shows the best performance on all problems. 
The final average HV and average EU scores of TensorRVEA in the four environments surpass those of NSGA-II and random search. 
However, it is important to note that the confidence intervals for TensorRVEA tend to increase during the later phases, suggesting a potential variability in its performance outcomes.
In contrast, the confidence intervals for both NSGA-II and random search remain comparatively narrow. 
The performance of NSGA-II, as measured by HV in the MoHalfCheetah and MoSwimmer environments, is inferior to that achieved by random search.
This observation points to the varying effectiveness of these algorithms across different problem settings. 
In the MoHopper-m3 environment, NSGA-II and RS show similar performance in the HV and the EU metrics. 
This resemblance could be attributed to the dominance resistance phenomenon and a swift increase in nondominated solutions.
TensorRVEA consistently yields high-quality and diverse policies across all tested environments, underlining its versatility. 
This superior performance is attributable to its selection mechanism based on reference vectors. Unlike the non-dominated sorting method, this method facilitates evolution in the direction of the reference vectors, enabling it to effectively escape local optima.
Notably, diversity is a crucial factor in robotic control tasks, catering to varied requirements such as high-speed operation at the expense of energy efficiency or the converse. 
In the MoSwimmer environment, NSGA-II produces more evenly distributed solutions, yet there remains room for improvement in solution quality. 
Meanwhile, the solutions generated by random search tend to lose diversity, with multiple solutions emerging at specific performance levels.

\subsection{Extensibility}
In this experiment, the extensibility of TensorRVEA was tested by employing a variety of reproduction operators.
Within the framework of TensorRVEA, we integrated several representative reproduction operators, including Genetic Algorithm (GA)~\cite{ga}, Particle Swarm Optimization (PSO)~\cite{pso}, Differential Evolution (DE)~\cite{de}, Competitive Swarm Optimizer (CSO)~\cite{cso}, as well as the Random reproduction serving as the baseline for comparison. 
Here, CSO is picked, as it is tailored for large-scale optimization, a required feature for neuroevolution.
The population size was fixed at \num{10000}, and each algorithm was executed over 100 generations in both the 2-objective and 3-objective MoHopper environments.
The performance of these variations was evaluated in terms of HV and EU.

The results are presented in Figure \ref{fig:RL2}.
Notably, CSO demonstrates the best performance in both two-objective and three-objective environments. 
This enhanced performance can be attributed to the feature that CSO is tailored for large-scale optimization. 
An intriguing observation is that PSO and DE achieved similar convergence rates to CSO in the initial phase.
However, the convergence of both PSO and DE seems to get stuck in the later phase. 
In contrast, the performance of GA seems well-balanced, outperforming DE and PSO, though not as effective as CSO.

\section{Conclusion}
For advancing evolutionary multiobjective optimization (EMO) towards addressing massive-scale and many-objective problems, our work demonstrates the unique advancements of tensorization and GPU acceleration, using TensorRVEA as the instance.
The robustness and efficiency of TensorRVEA is assessed on a diverse range of optimization challenges, including traditional numerical benchmark and high-dimensional neuroevolution challenges for robotic control tasks. 
The experimental results highlight TensorRVEA's notable computational speed and enhanced performance, which is attributed to its fully tensorized design and GPU-accelerated implementation. 
Furthermore, the algorithm's adaptability in integrating various reproduction operators further underscores its broad extensibility. 
Looking ahead, we will further explore the tensorization of general EMO algorithms, probably towards the development of tailored algorithm frameworks and operators for GPU acceleration.



\bibliographystyle{ACM-Reference-Format}
\bibliography{sample-base}

\appendix

\onecolumn
\section*{\centering\huge Supplementary Material}
\addcontentsline{toc}{section}{Supplementary Material}

\section{Main framework of RVEA}
\label{appendix:rvea}

\begin{algorithm}
    \caption{Main Framework of RVEA}
    \label{algo:rvea-pseudocode}
    
    \begin{algorithmic}[1]
        \State \textbf{Input:} Population size \( n \), the number of reference vectors \(r\), maximal number of generations \( t_{\text{max}} \), unit reference vectors \( V_0 = \{\mathbf{v}_{0,1}, \mathbf{v}_{0,2}, \ldots, \mathbf{v}_{0,r}\} \)
        \State \textbf{Output:} Final population \( P_{t_{\text{max}}} \)
        \State \textbf{Initialization}: Initialize population \( P_0 \) with \( n \) randomized individuals
        \For{\( t = 1 \) to \( t_{\text{max}} \)}
            \State Generate offspring \( Q_t \) from \( P_t \);
            \State Combine parent and offspring populations to create \( P_t \);
            \State Perform RVEA-Selection on \( P_t \) using \( V_t \) to obtain \( P_{t+1} \);
            \State Update reference vectors to obtain \( V_{t+1} \) from \( V_t \) and \( V_0 \);
        \EndFor
    \end{algorithmic}
\end{algorithm}

\section{Tensorized Crossover and Mutation}
\label{appendix:cross_mut}

Given a tensorized population $\boldsymbol{X} \in \mathbb{R}^{n \times d}$, it is partitioned into two tensorized parent populations $\boldsymbol{X}_1, \boldsymbol{X}_2 \in \mathbb{R}^{\lfloor \frac{n}{2} \rfloor \times d}$ for crossover. Random tensors $\boldsymbol{M}_c, \boldsymbol{R}_1, \boldsymbol{R}_2 \in [0, 1]^{\lfloor \frac{n}{2} \rfloor \times d}$ and $\boldsymbol{R}_3 \in [0, 1]^{\lfloor \frac{n}{2} \rfloor \times 1}$ are generated to facilitate the crossover process. The tensorized SBX can be represented using the following equations:

\begin{equation}
	\boldsymbol{B} = \text{sgn}(\boldsymbol{R}_1 - \frac{1}{2}) \cdot \Bigg[H(\frac{1}{2} - \boldsymbol{M}_c) \odot (2 \cdot \boldsymbol{M}_c)^{\frac{1}{\eta + 1}} +  (1 - H(\frac{1}{2} - \boldsymbol{M}_c)) \odot (2 - 2 \cdot \boldsymbol{M}_c)^{-\frac{1}{\eta + 1}} \Bigg],
\end{equation}
\begin{equation}
	\boldsymbol{B}  =\left[1- \text{repeat}(H(\boldsymbol{R}_3 - p_c), d)\right] \odot \left[ (1 - H(\boldsymbol{R}_2 - \frac{1}{2})) \odot \boldsymbol{B} + \right. \left. H(\boldsymbol{R}_2 - \frac{1}{2}) \right] +\text{repeat}(H(\boldsymbol{R}_3 - p_c), d),
\end{equation}
where $\eta$ is distribution parameter of SBX, $\text{sgn}(\cdot)$ returns 1 for elements $\geq 0$ and -1 for elements $< 0$, $H(\cdot)$ denotes a step function, $\text{repeat}(\boldsymbol{R}_3 > p_c, d)$ replicates the boolean tensor $\boldsymbol{R}_3 > p_c$ along the decision variable dimension $d$ and $p_c$ is the probability of crossover. Tensors $\boldsymbol{B}$ is an intermediate tensor used in the crossover calculation.

The offspring tensor $\boldsymbol{X}_{\text{cross}}$ is then calculated as:
\begin{equation}
	\boldsymbol{X}_{\text{cross}} = \begin{bmatrix}
		[(1 + \boldsymbol{B}) \odot \boldsymbol{X}_1 + (1 - \boldsymbol{B}) \odot \boldsymbol{X}_2]/2 \\ 
		[(1 - \boldsymbol{B}) \odot \boldsymbol{X}_1 + (1 + \boldsymbol{B}) \odot \boldsymbol{X}_2]/2
	\end{bmatrix},
\end{equation}
where the operation concatenates the two tensors along the first dimension, effectively combining the offspring tensors from the two parent sets.

Polynomial Mutation introduces subtle modifications to the offspring generated by crossover in TensorRVEA. This mutation process is defined by randomly generated tensor $\boldsymbol{R}_4 \in [0, 1]^{n \times d}$.The probability of a mutation occurring is denoted by $p_m$, and the magnitude of the mutation is represented by the tensor $\boldsymbol{M}_\text{mut}$. The tensorized polynomial mutation process is mathematically represented as follows:

\begin{equation}
	\boldsymbol{C} = H(p_m/d - \boldsymbol{R}_4),
\end{equation}

\begin{equation}
    \boldsymbol{X}_{\text{mut}} = \boldsymbol{X}_{\text{cross}} + \Delta \odot \boldsymbol{C},
\end{equation}

\begin{equation}
    \begin{split}
        \Delta &= (\boldsymbol{U} - \boldsymbol{L}) \cdot \Bigg[ \Bigg(2 \cdot \boldsymbol{M}_\text{mut} + (1 - 2 \cdot \boldsymbol{M}_\text{mut}) \cdot \left( 1 - \frac{\boldsymbol{X}_{\text{cross}} - \boldsymbol{L}}{\boldsymbol{U} - \boldsymbol{L}} \right)^{\xi + 1} \Bigg)^{\frac{1}{\xi + 1}} - 1 \Bigg] \cdot H(0.5 - \boldsymbol{M}_\text{mut}) \\
        &\qquad + (\boldsymbol{U} - \boldsymbol{L}) \cdot \Bigg[ 1 - \Bigg(2 \cdot (1 - \boldsymbol{M}_\text{mut}) + 2 \cdot (\boldsymbol{M}_\text{mut} - 0.5) \cdot \left( 1 - \frac{\boldsymbol{U} - \boldsymbol{X}_{\text{cross}}}{\boldsymbol{U} - \boldsymbol{L}} \right)^{\xi + 1} \Bigg)^{\frac{1}{\xi + 1}} \Bigg] \cdot H(\boldsymbol{M}_\text{mut} - 0.5),
    \end{split}
\end{equation}
where $\boldsymbol{C}$ is a tensor that determines the likelihood of mutation at each site. The tensors $\boldsymbol{U}$ and $\boldsymbol{L}$, representing the upper and lower bounds of the decision variable, are replicated $n$ times.
$\xi$ is distribution parameter of mutation.

\section{Experiments}
\label{appendix:exp_all}

\subsection{Acceleration Performance}
The experimental setups for Series 1 and Series 2 within the acceleration performance study are detailed in Tables \ref{tab:experiment_setup_series1} and \ref{tab:experiment_setup_series2}, respectively.

\begin{table}[h]
    \centering
    \caption{Experimental setup for Series 1: varying population size}
    \label{tab:experiment_setup_series1}
    \begin{tabular}{@{}cc@{}}
    \toprule
    \textbf{Aspect}              & \textbf{Details}                                             \\ \midrule
    \textbf{Problem}             &  DTLZ1                   \\ 
    \textbf{Algorithms}          & RVEA (CPU), TensorRVEA (CPU), RVEA (GPU), TensorRVEA (GPU)   \\
    \textbf{Repetitions}         & 10 independent iterations per algorithm                     \\
    \textbf{Generations}         & 100                                                          \\ 
    \textbf{Metrics}             & Average computation time per generation                      \\
    \textbf{Decision Dimension}  & \( d = 100 \)                                                \\
    \textbf{Number of Objectives}& \( m = 3 \)                                                 \\
    \textbf{Population Size}     & 32, 64, 128, 256, 512, \num{1024}, \num{2048}, \num{4096}, \num{8192}, \num{16384}         \\ \bottomrule
    \end{tabular}
\end{table}

\begin{table}[h]
    \centering
    \caption{Experimental setup for Series 2: varying decision dimension}
    \label{tab:experiment_setup_series2}
    \begin{tabular}{@{}cc@{}}
    \toprule
    \textbf{Aspect}              & \textbf{Details}                                             \\ \midrule
    \textbf{Problem}           &  DTLZ1                   \\ 
    \textbf{Algorithms}          & RVEA (CPU), TensorRVEA (CPU), RVEA (GPU), TensorRVEA (GPU)   \\
    \textbf{Repetitions}         & 10 independent iterations per algorithm                     \\
    \textbf{Generations}         & 100                                                          \\ 
    \textbf{Metrics}             & Average computation time per generation                      \\
    \textbf{Population Size}     & \( n = 100 \)                                                \\
    \textbf{Number of Objectives}& \( m = 3 \)                                                 \\
    \textbf{Decision Dimension}  & 512, \num{1024}, \num{2048}, \num{4096}, \num{8192}, \num{16384}, \num{32768}, \num{65536}, \num{131072}, \num{262144} \\ \bottomrule
    \end{tabular}
\end{table}

\subsection{Numerical benchmarks}
The experimental setups of numerical benchmarks experiments are shown in Table \ref{tab:numerical_benchmarks_setup}. For a comprehensive analysis, two key performance indicators are employed: Inverted Generational Distance (IGD)~\cite{igd} and Hypervolume (HV)~\cite{hv}. The IGD is defined as:

\begin{equation}
    \text{IGD}(\boldsymbol{F}, \boldsymbol{F}^*) = \frac{\sum_{\boldsymbol{f}^* \in \boldsymbol{F}^*} \min_{\boldsymbol{f} \in \boldsymbol{F}} ||\boldsymbol{f} - \boldsymbol{f}^*||}{|\boldsymbol{F}^*|}
\end{equation}
where \( \boldsymbol{F} \) represents the final solutions, \( \boldsymbol{F}^* \) is the reference Pareto Front, and \( ||\cdot|| \) denotes the Euclidean distance.

The HV is given by:
\begin{equation}
    \label{eq:HV_calculation}
    \text{HV}(\boldsymbol{F}, \mathbf{v}_{\text{ref}}) = \bigcup_{\boldsymbol{f} \in \boldsymbol{F}} \text{volume}(\mathbf{v}_{\text{ref}}, \boldsymbol{f}),
\end{equation}
where \(\text{volume}(\mathbf{v}_{\text{ref}}, \boldsymbol{f})\) denotes the volume of the hypercube defined by the vector \( \boldsymbol{f} \) and the reference point \( \mathbf{v}_{\text{ref}} \). In this context, \( \boldsymbol{F} \) represents the set of normalized objective values, and the reference point \( \mathbf{v}_{\text{ref}} \) is predetermined to be a vector of ones.

\begin{figure*}[h]
    \centering

    \begin{subfigure}[b]{0.24\linewidth}
        \includegraphics[width=\linewidth]{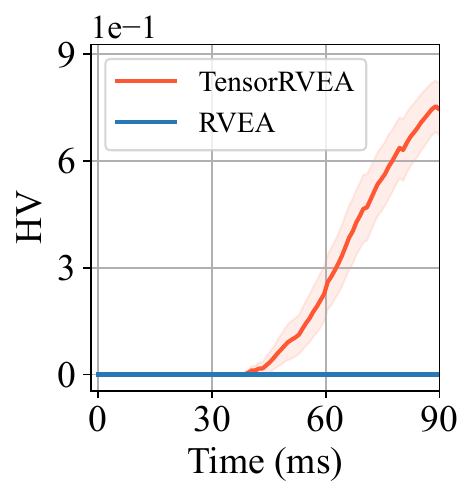}
		\captionsetup{skip=-2pt}
        \caption{DTLZ1}
        \label{fig:DTLZ1_HV}
    \end{subfigure}
    \hfill
	\begin{subfigure}[b]{0.24\linewidth}
        \includegraphics[width=\linewidth]{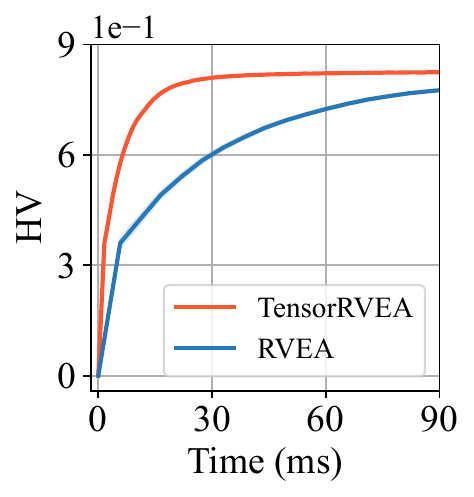}
		\captionsetup{skip=-2pt}
        \caption{DTLZ2}
        \label{fig:DTLZ2_HV}
    \end{subfigure}
    \hfill
    \begin{subfigure}[b]{0.24\linewidth}
        \includegraphics[width=\linewidth]{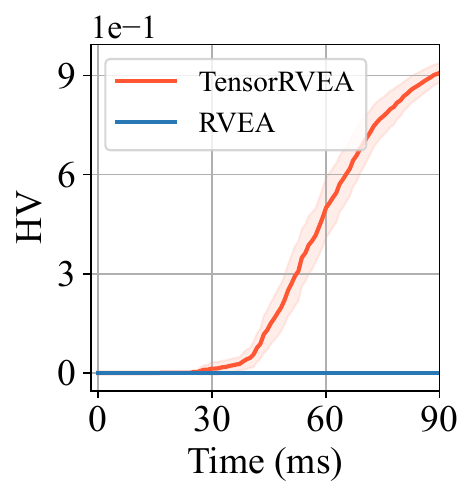}
		\captionsetup{skip=-2pt}
        \caption{DTLZ3}
        \label{fig:DTLZ3_HV}
    \end{subfigure}
    \hfill
	\begin{subfigure}[b]{0.24\linewidth}
        \includegraphics[width=\linewidth]{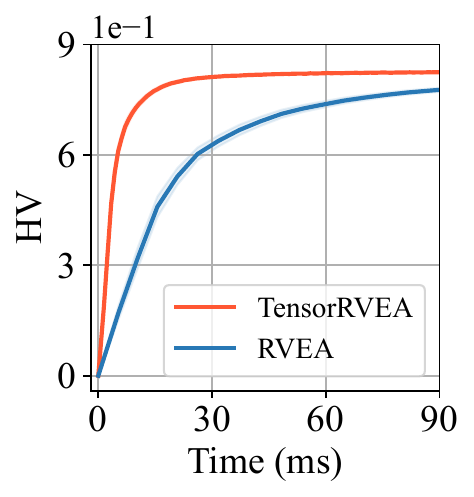}
		\captionsetup{skip=-2pt}
        \caption{DTLZ4}
        \label{fig:DTLZ4_HV}
    \end{subfigure}

    \captionsetup{skip=5pt}
    \caption{Comparison of mean HV values and 95\% confidence intervals for TensorRVEA and RVEA on DTLZ1-DTLZ4 problems.}
    \label{fig:all_DTLZ_HV}
\end{figure*}

Figure \ref{fig:all_DTLZ_HV} shows the HV curves of TensorRVEA and RVEA algorithms. The results demonstrate that TensorRVEA presents rapid convergence, achieving this within \SI{90}{\milli\second}. Notably, in the case of the DTLZ1 and DTLZ3 problems, the HV indicator for RVEA consistently registered a value of zero.

\begin{table}[h]
    \centering
    \caption{Experimental setup for conducting numerical benchmarks on DTLZ problems}
    \label{tab:numerical_benchmarks_setup}
    \begin{tabular}{@{}cc@{}}
    \toprule
    \textbf{Aspect}                 & \textbf{Details}                                              \\ \midrule
    \textbf{Algorithms}             & TensorRVEA, RVEA                                              \\
    \textbf{Problems and Dimensions} & DTLZ1 (\(d=7\)), DTLZ2 (\(d=12\)), DTLZ3 (\(d=12\)), DTLZ4 (\(d=12\)) \\
    \textbf{Performance Metric}     & Inverted Generational Distance (IGD) , Hypervolume (HV)                         \\
    \textbf{Repetitions}            & 31 independent runs                                          \\
    \textbf{Population Size}        & 105                                                          \\
    \textbf{Algorithm Stop Time}    & \SI{90}{\milli\second}                                                         \\
    \bottomrule
    \end{tabular}
\end{table}

\subsection{Multiobjective Neuroevolution}
\label{appendix:mone}

\begin{figure}[ht]
    \centering
    \begin{subfigure}{.32\linewidth}
        \includegraphics[width=\linewidth]{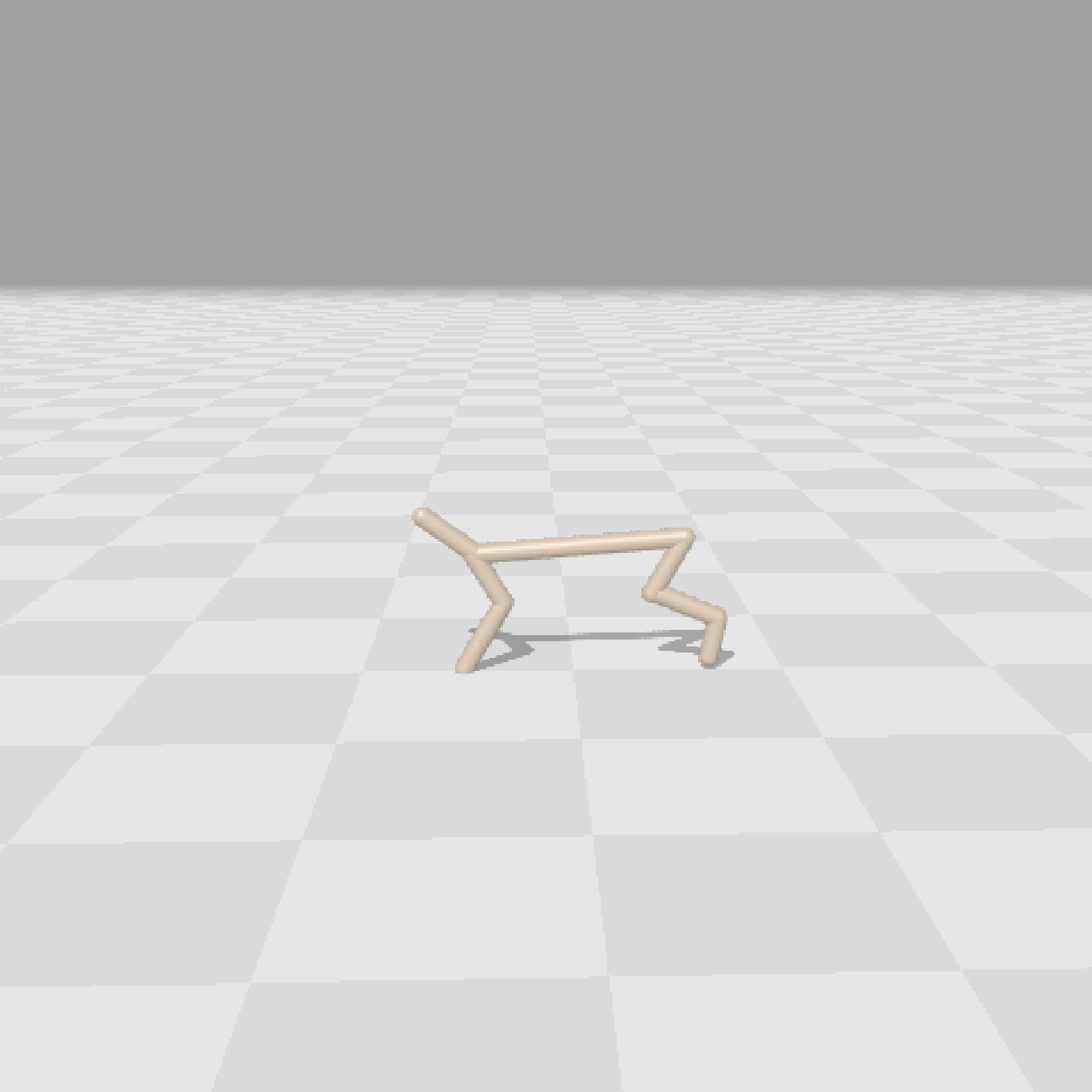}
        \caption{MoHalfCheetah}
        \label{fig:halfcheetah_fig}
    \end{subfigure}
    \begin{subfigure}{.32\linewidth}
        \includegraphics[width=\linewidth]{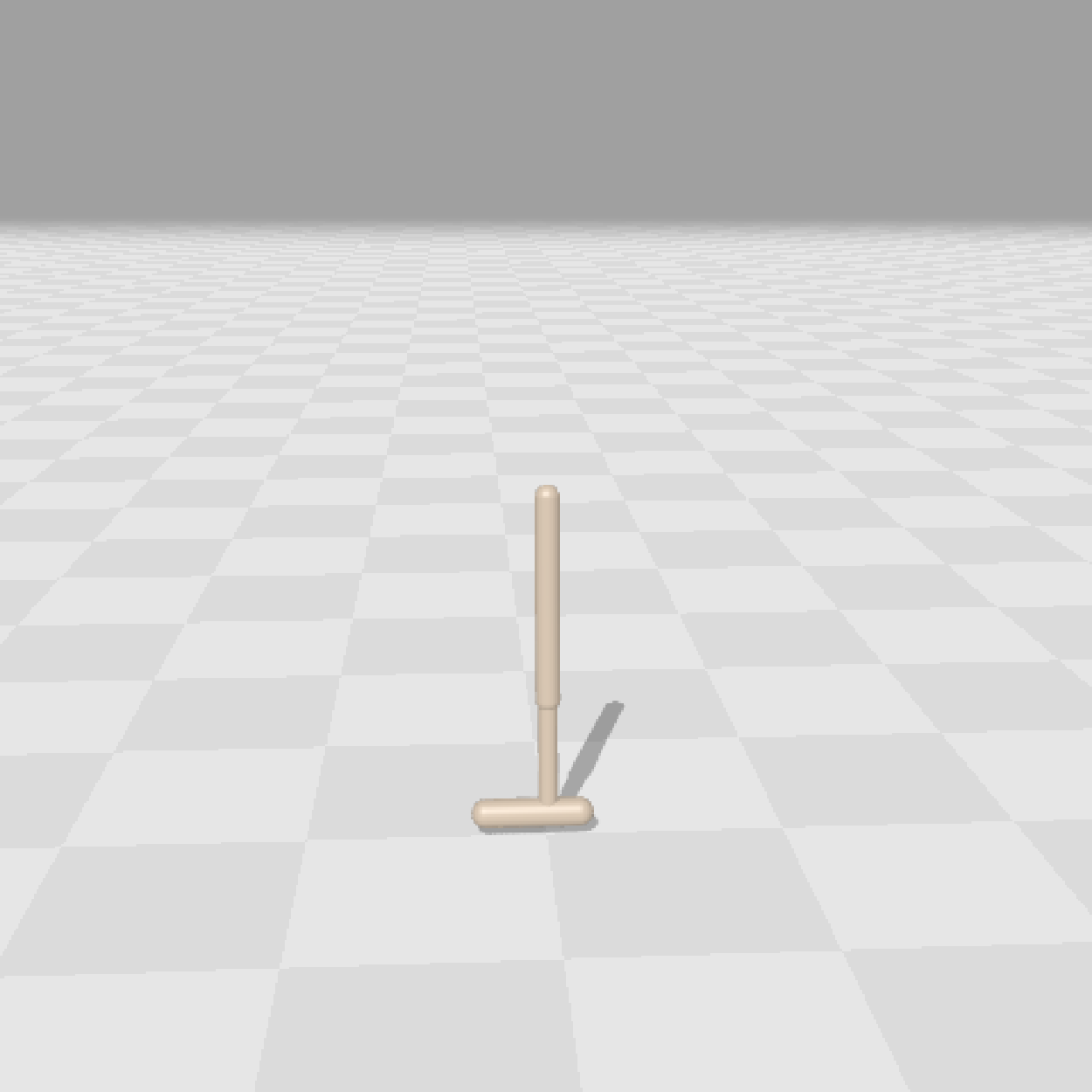}
        \caption{MoHopper}
        \label{fig:hopper_fig}
    \end{subfigure}
    \begin{subfigure}{.32\linewidth}
        \includegraphics[width=\linewidth]{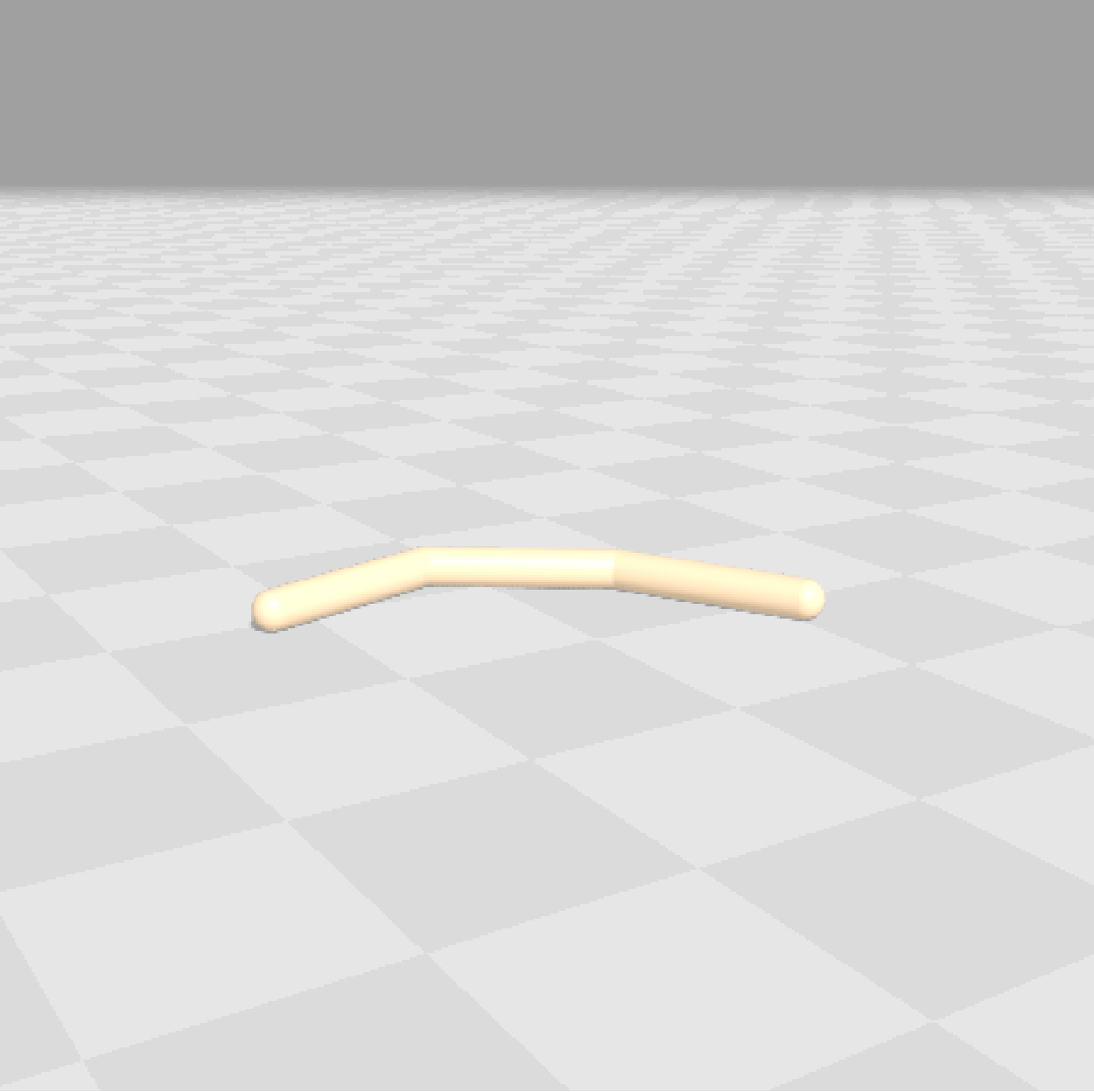}
        \caption{MoSwimmer}
        \label{fig:swimmer_fig}
    \end{subfigure}
    \caption{Robotic control tasks involved in the experiment for multiobjective neuroevolution.}
    \label{fig:brax_fig}
\end{figure}

In this section, we provide a comprehensive overview of the robotic control tasks tailored for multiobjective neuroevolution. We designed 4 multiobjective optimization problems based on three environments in Brax~\cite{brax}. 
The visualization of these three environments is shown in Figure \ref{fig:brax_fig}. 
In this context, the state space, denoted by \( \mathcal{S} \), encapsulates all possible configurations and statuses that the robotic system can attain. Conversely, the action space, represented by \( \mathcal{A} \), comprises the set of all actionable controls or decisions that the robot can execute to transition from one state to another within the environment. 

\subsubsection{Definition of Multiobjective Neuroevolution Problems}
The specific definitions of the four problems are as follows.

\textbf{MoHalfCheetah}:
The dimensionality of the observation space and the action space are denoted as \( \mathcal{S} \in \mathbb{R}^{17} \) and \( \mathcal{A} \in \mathbb{R}^{6} \), respectively. Each episode is comprised of 1000 steps.
The first objective is forward reward:
\begin{equation}
    f_1 =  w_1 \cdot v_x,
\end{equation}
where \( v_x\) is the velocity in \( x\) direction and \( w_1\) is weight for the velocity.
The second objective is control cost:
\begin{equation}
    f_2 = - w_2 \cdot \sum_i{{a_i}^2},
\end{equation}
where \( a_i\) is the action of each actuator and \( w_2\) is weight for the control cost.

\textbf{MoHopper-m2}:
The dimensionality of the observation space and the action space are denoted as \( \mathcal{S} \in \mathbb{R}^{11} \) and \( \mathcal{A} \in \mathbb{R}^{3} \), respectively. Each episode is comprised of 1000 steps.
The first objective is forward reward:
\begin{equation}
    f_1 =  w_1 \cdot v_x - \sum_i{{a_i}^2} + C,
\end{equation}
where \( v_x \) represents the velocity in the \( x \)-direction, \( w_1 \) is the weight  for this velocity component, \( a_i \) is the action of each actuator, and \( C = 1 \) constitutes a survival reward, indicating that the agent remains alive.
The second objective is height:
\begin{equation}
    f_2 = 10 \cdot (h_\text{curr} - h_\text{init}) - \sum_i{{a_i}^2} + C,
\end{equation}
where \( h_\text{curr}\) is the current height of the hopper and \( h_\text{init}\) is the initial height.

\textbf{MoHopper-m3}:
The dimensionality of the observation space and the action space are denoted as \( \mathcal{S} \in \mathbb{R}^{11} \) and \( \mathcal{A} \in \mathbb{R}^{3} \), respectively. Each episode is comprised of 1000 steps.
The first objective is forward reward:
\begin{equation}
    f_1 =  w_1 \cdot v_x + C,
\end{equation}
where \( v_x \) represents the velocity in the \( x \)-direction, \( w_1 \) is the weight  for this velocity component, and \( C = 1 \) constitutes a survival reward, indicating that the agent remains alive.
The second objective is height:
\begin{equation}
    f_2 = 10 \cdot (h_\text{curr} - h_\text{init})+ C,
\end{equation}
where \( h_\text{curr}\) is the current height of the hopper and \( h_\text{init}\) is the initial height.
The third objective is control cost:
\begin{equation}
    f_3 =  - \sum_i{{a_i}^2} + C, 
\end{equation}
where \( a_i \) is the action of each actuator.

\textbf{MoSwimmer}:
The dimensionality of the observation space and the action space are denoted as \( \mathcal{S} \in \mathbb{R}^{8} \) and \( \mathcal{A} \in \mathbb{R}^{2} \), respectively. Each episode is comprised of 1000 steps.
The first objective is forward reward:
\begin{equation}
    f_1 =  w_1 \cdot v_x,
\end{equation}
where \( v_x\) is the velocity in \( x\) direction and \( w_1\) is weight for the velocity.
The second objective is control cost:
\begin{equation}
    f_2 = - w_2 \cdot \sum_i{{a_i}^2},
\end{equation}
where \( a_i\) is the action of each actuator and \( w_2\) is weight for the control cost.

\subsubsection{Experimental Setup}
The experimental setups of multiobjective neuroevolution experiments are shown in Table \ref{tab:multiobjective_neuroevolution_setup}. 
\begin{table}[htbp]
    \centering
    \caption{Experimental setup for multiobjective neuroevolution in robotic control}
    \label{tab:multiobjective_neuroevolution_setup}
    \begin{tabular}{@{}cc@{}}
    \toprule
    \textbf{Aspect}                      & \textbf{Details}                                             \\ \midrule
    \textbf{Problems}     & MoHalfCheetah, MoHopper-m2, \\
                                             & MoHopper-m3, MoSwimmer\\
    \textbf{Objectives and Dimensions}                  & Varies (see Table \ref{tab:environment-objectives})           \\
    \textbf{Algorithms}                  & TensorRVEA, NSGA-II, Random Search                            \\
    \textbf{Policy Model}              & MLP with 1 hidden layer (16 neurons) and Tanh activation function                 \\
    \textbf{Repetitions}                 & 10 independent runs per algorithm                            \\
    \textbf{Population Size}             & \num{10000}                                                       \\
    \textbf{Stop Time}            & MoHalfCheetah: \SI{15000}{\second}, MoHopper-m2: \SI{400}{\second}, \\
                                             & MoSwimmer: \SI{800}{\second}, MoHopper-m3: \SI{400}{\second}                            \\
    \textbf{Performance Metrics}         & Hypervolume (HV), Expected Utility (EU), Visualizations      \\
    \bottomrule
    \end{tabular}
\end{table}

The Expected Utility (EU)~\cite{eu} metric is calculated by:
\begin{equation}
    \text{EU}(\boldsymbol{F}) = \mathbb{E}_{\mathbf{w} \sim \mathbf{W}} \left[ \max_{\boldsymbol{f} \in \boldsymbol{F}} \boldsymbol{f} \cdot \mathbf{w} \right],
\end{equation}
where \( \boldsymbol{F} \) denotes the final set of solutions obtained from the optimization process, and \( \mathbf{W} \) represents a probability distribution over the reward weights.

In the computation of the Hypervolume (HV), as detailed in Eq.~\ref{eq:HV_calculation}, the reference point is established by selecting the minimum objective values from the Pareto-optimal solutions obtained in all iterations of the TensorRVEA, NSGA-II, and random search algorithms.
If the minimum values of the objectives, particularly forward reward and height, are less than zero, the reference points for these objectives are set at zero. 
Table~\ref{tab:env_ref_points_1} details the reference points for each environment used in the HV computations.

\begin{table}[h!]
    \centering
    \caption{Reference points for HV calculation of multiobjective neuroevolution.}
    \label{tab:env_ref_points_1}
    \begin{tabular}{@{}cc@{}}
    \toprule
    Problem & Reference Point \\ \midrule
    MoHalfCheetah & (0, $-$599.78643799) \\
    MoHopper-m2 & (0, $-$865.70227051) \\
    MoSwimmer & (0, $-$0.19898804) \\
    MoHopper-m3 & (0, 0, $-$1942.84301758) \\ \bottomrule
    \end{tabular}
\end{table}

\subsection{Extensibility}
The experimental setups of the extensibility experiments are detailed in Table~\ref{tab:extensibility_setup}, while Table~\ref{tab:env_ref_points_2} presents the reference points established for each environment in the computations. These reference points are derived by selecting the minimal objective values from the Pareto-optimal solutions, accumulated over all iterations of the five variants of the TensorRVEA algorithm. In cases where the minimal values of certain objectives, such as forward reward and height, fall below zero, the reference points for these objectives are set to zero.

\begin{table}[htbp]
    \centering
    \caption{Experimental setup for testing the extensibility of TensorRVEA}
    \label{tab:extensibility_setup}
    \begin{tabular}{@{}cc@{}}
    \toprule
    \textbf{Aspect}                      & \textbf{Details}                                             \\ \midrule
    \textbf{Reproduction Operators}      & GA, PSO, DE, CSO, Random reproduction (baseline)               \\
    \textbf{Policy Model}              & MLP with 1 hidden layer (16 neurons) and Tanh activation function                 \\
    \textbf{Environments}                & MoHopper-m2 and MoHopper-m3                         \\
    \textbf{Population Size}             & \num{10000}                                                       \\
    \textbf{Generations}                 & 100                                                          \\
    \textbf{Performance Metrics}         & Hypervolume (HV), Expected Utility(EU), Visualizations                     \\
    \bottomrule
    \end{tabular}
\end{table}

\begin{table}[h!]
    \centering
    \caption{Reference points for HV calculation of extensibility.}
    \label{tab:env_ref_points_2}
    \begin{tabular}{@{}cc@{}}
    \toprule
    Environment & Reference Point \\ \midrule
    MoHopper-m2 & (0, $-$1127.1895752) \\
    MoHopper-m3 & (80.86401367, 0, $-$1905.52331543) \\ \bottomrule
    \end{tabular}
\end{table}

\end{document}